\documentclass[10pt,journal,compsoc]{IEEEtran}

\ifCLASSOPTIONcompsoc
  \usepackage[nocompress]{cite}
\else
  \usepackage{cite}
\fi

\usepackage[table,xcdraw]{xcolor}

\usepackage{graphicx}

\usepackage{tikz}
\usepackage{comment}
\usepackage{amsmath,amssymb} % define this before the line numbering.
\usepackage[colorlinks,linkcolor=red]{hyperref}
\usepackage{multirow}
\usepackage{multirow}
\usepackage{algorithm}  
\usepackage{algorithmicx}  
\usepackage{algpseudocode}  
\usepackage{amsmath}  
\usepackage{array}

\begin{document}

\title{Revisiting the Trade-off between Accuracy and Robustness via Weight Distribution of Filters}

\author{Xingxing~Wei,~\IEEEmembership{Member,~IEEE},~Shiji~Zhao,~and~Bo~Li$^*$ 
\IEEEcompsocitemizethanks{\IEEEcompsocthanksitem Xingxing Wei, Shiji Zhao, and Bo Li were at the Institute of Artificial Intelligence, Beihang University, No.37, Xueyuan Road, Haidian District, Beijing,
100191, P.R. China. (E-mail: \{xxwei, zhaoshiji123, boli\}@buaa.edu.cn)
\protect\\
% E-mail: \{zhaoshiji123, xxwei, wangxizhe\}@buaa.edu.cn
\IEEEcompsocthanksitem Bo Li is the corresponding author.}% <-this % stops an unwanted space
}

% \markboth{IEEE TRANSACTIONS ON PATTERN ANALYSIS AND MACHINE INTELLIGENCE}%
% {Shell \MakeLowercase{\textit{et al.}}: Bare Demo of IEEEtran.cls for Computer Society Journals}

\markboth{IEEE TRANSACTIONS ON PATTERN ANALYSIS AND MACHINE INTELLIGENCE}%
{Shell \MakeLowercase{\textit{et al.}}: Bare Demo of IEEEtran.cls for Computer Society Journals}

\IEEEtitleabstractindextext{
\begin{abstract}
Adversarial attacks have been proven to be potential threats to Deep Neural Networks (DNNs), and many methods are proposed to defend against adversarial attacks. However, while enhancing the robustness, the accuracy for clean examples will decline to a certain extent, implying a trade-off existed between the accuracy and adversarial robustness. In this paper, to meet the trade-off problem, we theoretically explore the underlying reason for the difference of the filters’ weight distribution between standard-trained and robust-trained models and then argue that this is an intrinsic property for static neural networks, thus they are difficult to fundamentally improve the accuracy and adversarial robustness at the same time. Based on this analysis, we propose a sample-wise dynamic network architecture named Adversarial Weight-Varied Network (AW-Net), which focuses on dealing with clean and adversarial examples with a ``divide and rule" weight strategy. The AW-Net adaptively adjusts the network's weights based on regulation signals generated by an adversarial router, which is directly influenced by the input sample. Benefiting from the dynamic network architecture, clean and adversarial examples can be processed with different network weights, which provides the potential to enhance both accuracy and adversarial robustness. A series of experiments demonstrate that our AW-Net is architecture-friendly to handle both clean and adversarial examples and can achieve better trade-off performance than state-of-the-art robust models.
\end{abstract}

\begin{IEEEkeywords}
Adversarial Examples, Adversarial Robustness, Dynamic Network Structure, Accuracy-Robustness Trade-off.
\end{IEEEkeywords}}

\maketitle

\IEEEdisplaynontitleabstractindextext

% \fi
\IEEEpeerreviewmaketitle

\IEEEraisesectionheading{\section{Introduction}}
\label{sec:intro}
\IEEEPARstart{D}{ue} to impressive performance, Deep Neural Networks (DNNs) have recently been applied in various areas. However, DNNs can be misled by adversarial examples produced by adding imperceptible adversarial perturbations on clean examples \cite{szegedy2013intriguing,madry2017towards,wei2023physically}, which 
indicates that serious security issues exist in DNNs. To defend against adversarial examples, some researches are proposed to enhance the model's robustness. 
Adversarial Training \cite{madry2017towards} has  proved to be an effective way to enhance the robustness of models.
However, as a price of robustness,  {the accuracy for clean examples} will decline to a certain extent \cite{zhang2019theoretically},  {which greatly limits the application in practical scenarios}.

\begin{figure}[t]
  \centering
\includegraphics[width=0.43\textwidth]{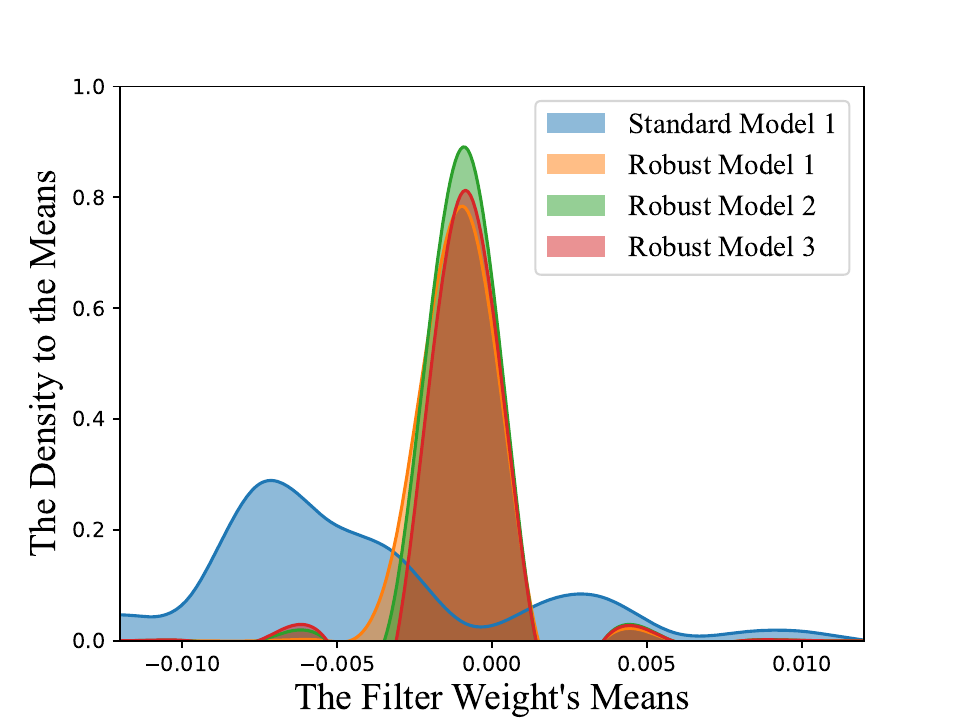}\\
\caption{The distribution versus the means of filters' weights in the second convolution layer in ResNet-18 on CIFAR-10 towards a standard model and three state-of-the-art robust models trained by \cite{wang2019improving,zi2021revisiting,Zhao2022Enhanced}. The figure shows filters' weight distribution exists an obvious difference for the standard and robust models.}
\label{distribution}
\end{figure}

\begin{figure*}[t]
  \centering
\includegraphics[width=0.9\textwidth]{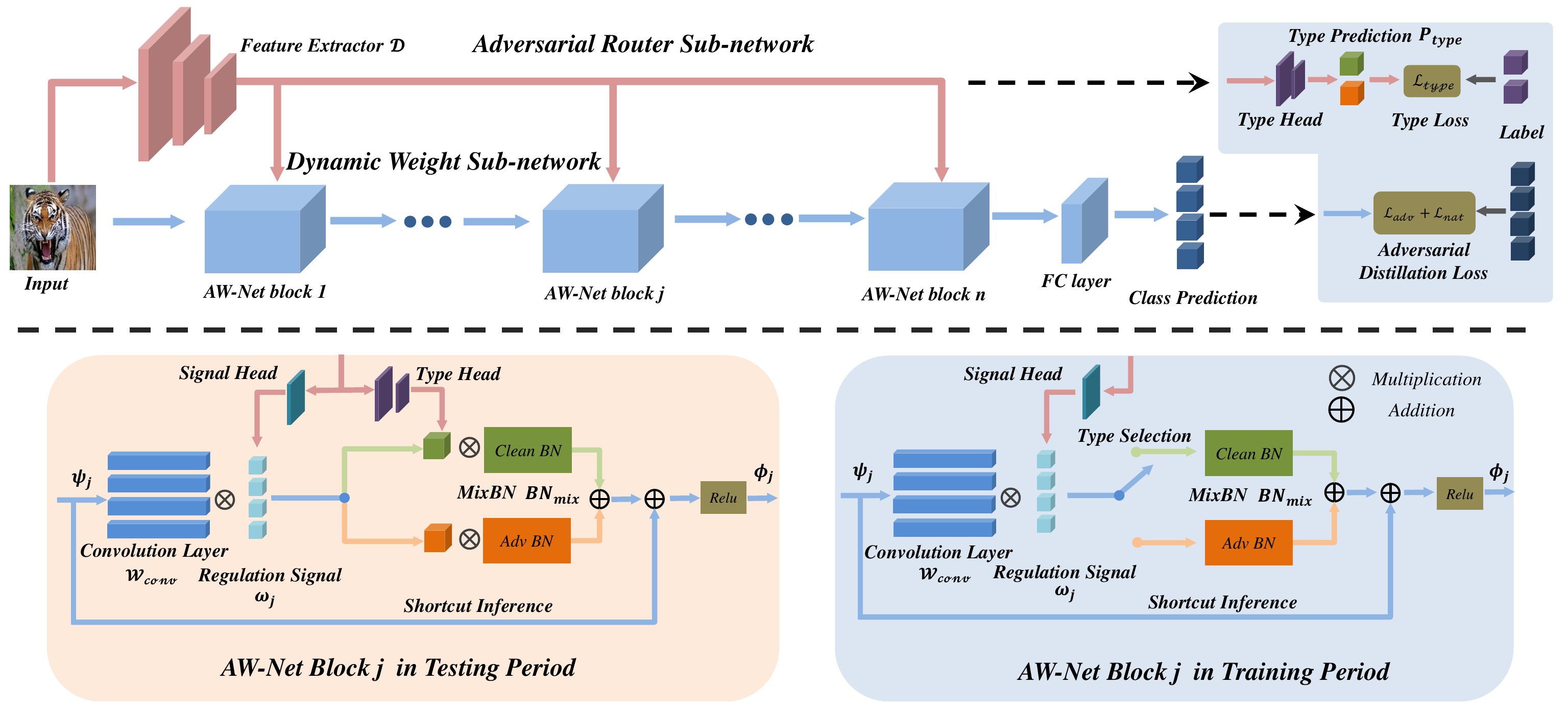}\\
\caption{The framework of our AW-Net. AW-Net includes two main branches: the dynamic weight sub-network composed of multiple AW-Net blocks and the adversarial router sub-network to discriminate the clean and adversarial examples and generate the regulation signals. In the training period, we utilize the MixBN (including Adv BN and Clean BN) to handle different feature distributions of clean and adversarial examples, respectively. In the testing period, Adv BN and Clean BN are weighted by the prediction of adversarial router. }
\label{fig:framework}
\end{figure*}

To balance the trade-off between accuracy and adversarial robustness, various methods are presented from different views, such as advanced optimization algorithm \cite{zhang2019theoretically,Zhao2022Enhanced} or modified network architecture \cite{zhang2020auxiliary,meng2017magnet,dong2022adversarially}, and so on.   {However, the above methods ignore the essential impact of weight parameters. In fact, after the optimization process, the model performance is actually determined by its weight parameters, which directly reflect the most essential characteristics of the model. Some studies \cite{galloway2018adversarial,langenberg2019effect,qin2019adversarial} explore the adversarial robustness of models from a weight perspective, but they do not discuss the trade-off phenomena. In order to further understand the trade-off between accuracy and adversarial robustness, it is necessary to analyze the weight parameters of the model and take corresponding measures to fundamentally mitigate this phenomenon.
}

In this paper, we investigate the standard and robust models from the view of networks' weight distribution. In Figure \ref{distribution}, we observe that filters' weight distribution for the same network architecture between standard-trained and robust-trained models exists an obvious distinction, while the robust-trained models have a similar distribution. We analyze the underlying reason and give a theoretical explanation for this phenomenon in terms of the gradient regularization of different optimization algorithms. The empirical observation and theoretical analysis indicate that a static network is challenging to eliminate the gap and is hard to improve both the accuracy and adversarial robustness because it is difficult to contain two groups of different weights within a static network. A reasonable solution is to adaptively change the network's weights based on the input samples, and thus utilize the optimal weights to process them, respectively.

Inspired by the dynamic networks \cite{han2021dynamic, wang2020deep,li2021dynamic}, we design the Adversarial Weight-Varied Network (AW-Net) to perform a “divide and rule” weight strategy for the clean and adversarial examples. Specifically, AW-Net contains two branches: a dynamic weight sub-network and an adversarial router sub-network. The dynamic weight sub-network is in charge of changing the network's weights at the level of filters to process the clean and adversarial examples, respectively.  To handle different characteristic distributions of input samples and keep the model stable during training and testing, a MixBN \cite{liu2020towards} structure is adapted into the sub-network. The adversarial router sub-network is responsible for classifying the clean and adversarial examples, and then producing the regulation signals to guide the weights' change in the dynamic weight sub-network. These two sub-networks are jointly trained in an end-to-end adversarial training manner and can interact with each other closely. With these careful designs, AW-Net has the potential to enhance the accuracy and adversarial robustness at the same. The flowchart is shown in Figure \ref{fig:framework}. Our code is available at \url{https://github.com/zhaoshiji123/AW-Net}.

We summarize our contributions as follows:
\begin{itemize}
\item  {We argue that static neural networks are difficult to fundamentally address the accuracy-robustness trade-off challenge. To support this viewpoint, we theoretically exploring the underlying reason for the observed difference of filters’ weight distribution between standard-trained and robust-trained models, and show that static neural networks have an intrinsic limitation because it is hard to contain two
groups of different weights in a static network.}
\item We further propose a novel sample-wise dynamic network named Adversarial Weight-Varied Network (AW-Net), which adaptively adjusts the weights based on the input samples. For that, a dynamic weight sub-network at the filter level is designed, and a MixBN structure is used to handle the feature distributions for clean and adversarial examples. Meanwhile, we apply a multi-head adversarial router sub-network to generate the regulation signal to guide the dynamic weights. 
\item We extensively verify the effectiveness of AW-Net on  {three} public datasets and compare them with state-of-the-art robust models against white-box and black-box attacks, respectively. Results show that our method achieves the best performance for the trade-off between accuracy and adversarial robustness.
\end{itemize}

The rest of the paper is organized as follows: we introduce the related work in Section \ref{relatedwork}. Section \ref{differences} explores the differences in the filters' weight distribution. We present the AW-Net in Section \ref{AGnet}. The experiments are conducted in Section \ref{experiments}, and the conclusion is given in  Section \ref{conclusion}.

\section{Related Work}
\label{relatedwork}

%-------------------------------------------------------------------------
\subsection{Adversarial Attacks}

Szegedy et al.\cite{szegedy2013intriguing} demonstrate that imperceptible adversarial perturbations on inputs can mislead the prediction of DNNs. After that, lots of adversarial attack methods are proposed, such as Fast Gradient Sign Method (FGSM)\cite{goodfellow2014explaining}, Projected Gradient Descent Attack (PGD) \cite{madry2017towards}, Carlini and Wagner Attack (C$\&$W) \cite{carlini2017towards}, and Jacobian-based Saliency Map Attack (JSMA) \cite{papernot2016limitations}. 
% One-pixel attack only uses one pixel to attack the image with pretty performance.
Generally speaking, adversarial attacks can be divided into white-box attacks \cite{goodfellow2014explaining,madry2017towards,carlini2017towards} and black-box attacks \cite{wei2022adversarial, wei2023efficient,wei2022simultaneously,wei2022sparse,wei2021black,wei2023adversarial}. White-box attacks usually generate adversarial examples based on the gradients of the target models. While black-box attacks perform attacks via the transfer-based strategy or query-based strategy according to how much information of the target model can be obtained.   {Recently, a self-adaptive adversarial attack method named AutoAttack (AA) \cite{croce2020reliable} is proposed, which consists of four attack methods, including Auto-PGD (APGD), Difference of Logits Ratio (DLR) attack, FAB-Attack \cite{croce2020minimally}, and the black-box Square Attack \cite{andriushchenko2020square}. AutoAttack has been a widely-used attack method.}

In this paper, we propose a novel network architecture, and then test its robustness against the strong white-box and black-box attacks, respectively.

%-------------------------------------------------------------------------

\subsection{Adversarial Robustness}

To defend against adversarial attacks, lots of research is proposed to enhance the robustness.
Adversarial Training \cite{madry2017towards,zhang2019theoretically,jia2022prior,jia2023improving} is a representative method among them. Madry et al. \cite{madry2017towards} formulate Adversarial Training as a min-max optimization. Further, Adversarial Robustness Distillation (ARD) methods  \cite{goldblum2020adversarially,Zhao2022Enhanced} use strong teacher models to guide the adversarial training process of small student models to enhance the robustness. 
Meanwhile, many pre-processing methods are applied to remove the effects of adversarial perturbations. For example, \cite{gu2014towards,bai2017alleviating} utilize the encoder structure to denoise the adversarial examples, which shows competitive performance against adversarial attacks. Image compression operations are also verified to be effective in dealing with adversarial perturbations, such as JPEG compression \cite{das2017keeping} and DCT compression \cite{akhtar2018defense}, etc.
In addition, re-normalizing the image from the adversarial version to the clean version \cite{shu2021encoding,dong2022adversarially} can enhance the model robustness from the view of distribution. Shu et al.\cite{shu2021encoding} use a batch normalization layer to achieve the clean feature by re-normalizing the adversarial feature with means and variances of clean features in adversarial training. Dong et al.\cite{dong2022adversarially} try to transform the adversarial feature into a clean distribution in object detection tasks. 

We can see that all the above methods aim to improve the  robustness of a static network. In contrast, our method designs a dynamic network to adaptively adjust the weights to process the clean and adversarial examples, respectively.

%  {
\subsection{Trade-off between Accuracy and Robustness}  {
The trade-off between accuracy and adversarial robustness has been widely studied \cite{zhang2019theoretically, zhang2020geometry,yang2020closer,pang2022robustness, Zhao2022Enhanced}. From the perspective of optimization algorithm, Zhang et al. \cite{zhang2019theoretically} apply the prediction of clean examples to guide the training process towards adversarial examples. Zhang et al. \cite{zhang2020geometry} enhance both accuracy and robustness by reweighting the examples. Stutz et al. \cite{stutz2019disentangling} claim that manifold analysis can be helpful in achieving accuracy and robustness. Yang et al. \cite{yang2020closer} argue that the trade-off can be mitigated by optimizing the local Lipschitz functions. Pang et al. \cite{pang2022robustness} apply local equivariance to describe the ideal behavior of robust models and facilitate the balance between accuracy and robustness. Zhao et al. \cite{Zhao2022Enhanced} apply multiple teacher models to improve both the accuracy and robustness of student models.}

 {
From the perspective of modifying network architecture, MagNet \cite{meng2017magnet} contains a separate detector network and a reformer network, which learns the clean distribution and further distinguishes adversarial examples. However, it is not robust to the adaptive attack \cite{carlini2017magnet}. AdvProp \cite{xie2020adversarial} uses an auxiliary Batch-Normalization (BN) layer to improve the robustness and can achieve good accuracy on clean examples. Furthermore, Liu et al. \cite{liu2020towards} propose multiple BN layers to control the domain-specific statistics for different attacks. Although they show some positive effects, these methods ignore solving the trade-off by analyzing the weights' effect for different types of models. }

 {In this paper, we attempt to understand accuracy and robustness from the perspective of model weight distributions and mitigate the trade-off issue with the architecture of dynamic network weights.
}

\subsection{Dynamic Networks}
Different from static networks, dynamic networks usually change their architectures based on the input samples \cite{han2021dynamic}. The purpose of designing dynamic networks is to reduce the cost of network inference and utilize the most effective sub-networks to get the correct prediction. For example,  Wang et al.\cite{wang2020deep} consider the filters as experts and select a part of them to activate some specific layers. Li et al. \cite{li2021dynamic} propose the Dynamic Slimmable Network to divide the inputs into easy and hard samples and then design a dynamic network to predict the samples' type.  Veit et al. \cite{veit2018convolutional} and Wang et al. \cite{wang2018skipnet} design to dynamically skip the residual structure and reduce the calculation cost. 

Recently, the adversarial robustness of dynamic networks has been evaluated \cite{haque2020ilfo,hong2021panda,hu2020triple}, where \cite{haque2020ilfo} is the first work to show the vulnerability of dynamic networks.  Hu et al.\cite{hu2020triple} argue that triple wins can be obtained for accuracy, robustness, and efficiency via a multi-exit dynamic network.  However, Hong et al. \cite{hong2021panda}  show that the efficiency of the multi-exit network can be reduced by the proposed slowdown attack. Transductive defenses \cite{wang2021fighting,wu2020adversarial} update the network's parameters in the testing process to dynamically defend against adversarial examples. 

We are different from the above works as follows: (1) we aim at defense rather than attacks like \cite{haque2020ilfo,hong2021panda}. (2) we focus on dynamically adjusting the network's weights rather than utilizing the multi-exit layers like \cite{hu2020triple}. The motivation of our method is based on the filter's weight observation in Figure \ref{distribution}, which has a different defense mechanism with \cite{hu2020triple}. (3) During the testing phase, our method does not need to update the model parameters, which is obviously different from transductive defenses \cite{wang2021fighting,wu2020adversarial}.
 
%-------------------------------------------------------------------------

\section{Filter Weight Distributions}
\label{differences}
\subsection{Empirical Observation}

\begin{figure*}[t]
\begin{center}
\includegraphics[width=0.99\textwidth]{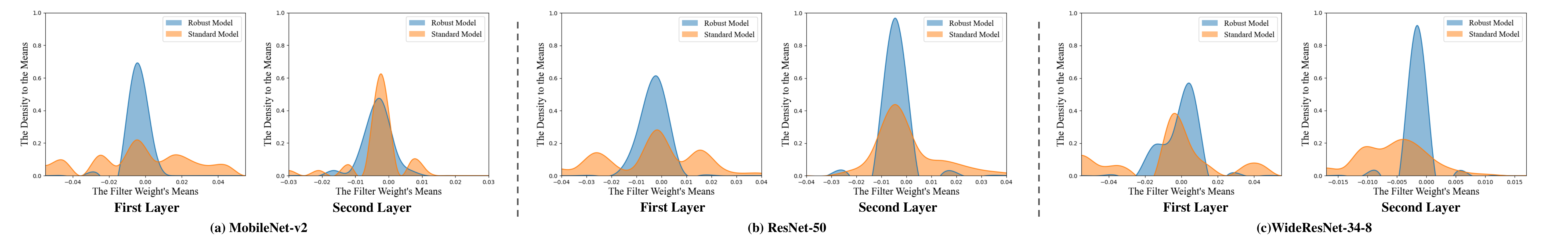}\\
% \begin{center}
\caption{The distribution versus the means of filters' weights in the first and second convolution layer in MobileNet-v2, ResNet-50, and WideResNet-34-8 on CIFAR-10 towards standard models and robust models trained by \cite{Zhao2022Enhanced}.  }
\label{fig:distribution_all}
\end{center}
\end{figure*}

%\begin{figure*}[th]
%  \centering
%\includegraphics[width=0.24\linewidth]{new_picture/layer1_cifar10_means.pdf}
%\includegraphics[width=0.24\linewidth]{new_picture/layer1_cifar10_vars.pdf}
%\includegraphics[width=0.24\linewidth]{new_picture/layer2_cifar10_means.pdf}
%\includegraphics[width=0.24\linewidth]{new_picture/layer2_cifar10_vars.pdf}
%\caption{The visualization of filters' means and variances for a strong standard model and a strong robust model trained by MART \cite{wang2019improving} on CIFAR-10. The left two figures are visualization of means and variances for the first layer in ResNet18; And the right two  figures are visualization of means and variances for the second layer in ResNet 18. For simplicity, we only show these two layers, and the other layers also show the same phenomenon.}
% \label{fig:result1}
%\end{figure*}
As we know, although some models have shown good adversarial robustness, the accuracy for clean examples is always inferior to the standard models. To investigate the underlying reason, we pay attention to the network's weights.  Because filters can be approximately considered as the independent units of feature extracting, we compute the means of the filters at different convolution layers and then use them to investigate the weight distributions. Figure \ref{distribution} shows that the means' distribution between the standard and robust models varies greatly. Specifically, the robust models have sharp distributions, while standard models have smooth distributions. To further explore this phenomenon, 
we test different network architectures. Here, besides the ResNet-18 network in Figure \ref{distribution}, we choose three different network structures including MobileNet-v2, ResNet-50, and WideResNet-34-8. The robust models are trained by MTARD \cite{Zhao2022Enhanced}. We visualize the weight distribution for the above three models and the result is shown in Figure \ref{fig:distribution_all}.
The results show that the phenomenon of weight gap widely exists in different network structures but is not just limited to a particular network structure, which can effectively support the existence of the model weights' gap between the standard model and the robust model. This evidence indicates that static neural networks are challenging to address the trade-off issue because it is hard to contain two groups of different weights in a static network, which motivates us to develop a novel neural network.

 {Some previous studies try to improve the adversarial robustness of DNN models from the perspective of weights \cite{langenberg2019effect,qin2019adversarial,galloway2018adversarial}, which seems similar to our work at first glance. Specifically, \cite{langenberg2019effect} has found that adversarial training actually can be regarded as a regularization that leads to low-rank and sparse weights for the trained models, and simultaneously low-rank and sparse weights can promote robustness against adversarial examples. \cite{qin2019adversarial} performs a Taylor expansion of the adversarial loss and then regularizes the high-order Taylor terms to obtain small weights, in this way, the adversarial robustness is improved. \cite{galloway2018adversarial} demonstrates that adversarial training can be seen as translating the data w.r.t. the loss, while weight decay scales the loss. The two approaches can be combined to yield one small model that has good adversarial robustness.  
}

 {However, there exist three major distinctions with ours as follows: (1) From the perspective of the target problem, these works aim at improving adversarial robustness while we aim at mitigating the trade-off between accuracy and adversarial robustness. The challenges of these two problems are different.  (2) From the perspective of research ideas, these works focus on exploring what are the intrinsic laws of filters' weights related to adversarial robustness. For that, they provide the empirical findings or theoretical analysis according to their viewpoints, such as weight decay in \cite{galloway2018adversarial} and low-rank weights in \cite{langenberg2019effect}. In contrast, we mainly want to show that static neural networks have the intrinsic limitation to deal with the trade-off issue, therefore, what causes the weights' distinction between standard-trained models and robust-trained models is our key concern, rather than the intrinsic laws of weights like \cite{langenberg2019effect,qin2019adversarial,galloway2018adversarial}.
Thus we give the theoretical proof to show this difference is inevitable for static neural networks (see Section 3.2). Because the angle of views is distinct, the consequent theoretical analysis between ours and \cite{langenberg2019effect,qin2019adversarial,galloway2018adversarial} are different, and we can see that the theoretical analysis in \cite{langenberg2019effect,qin2019adversarial,galloway2018adversarial} cannot support our viewpoint effectively. (3) From the perspective of the proposed methods, these works improve the adversarial robustness by upgrading optimization algorithms with additional regularization terms. Instead, owing to the intrinsic limitation of static neural networks to deal with the trade-off issue, we mitigate this challenge by designing a new dynamic network architecture, which is given in Section 4. In conclusion, our method has major differences with \cite{langenberg2019effect,qin2019adversarial,galloway2018adversarial}, and has our own merits. }

% all these methods do not analyze why the trade-off phenomenon occurs or provide corresponding solutions. }.

 %From the results,  we can see that the filter's weight of the standard model shows a rough change, while the filter's weight of the robust model shows a smooth change. This contrast illustrates the obvious difference between the standard and robust model. The  prominent means and variances of the standard model demonstrate that it is easy to capture more subtle image features, which can explain why the standard model is easily affected by adversarial samples with minor perturbations. As a comparison, the robust model shows smaller means and variances and is more robust to the adversarial perturbations. Different weight sizes between standard and robust models may directly lead to the sensitivity to clean and adversarial examples and can further influence the model's robustness.  

\subsection{Theoretical Analysis}
Inspired by \cite{simon2019first}, we theoretically analyze the above performance from the perspective of model's optimization, which directly influences the weight distribution. We investigate the intrinsic mechanisms of robust models using adversarial training methods based on PDG-AT \cite{madry2017towards} and TRADES \cite{zhang2019theoretically} following \cite{simon2019first,qin2019adversarial}. Firstly, we analyze the optimization loss of standard models and adversarial models during the training process. The optimization loss $L_{adv}(x,y)$ for adversarial training  and the optimization loss  $L(x,y)$ for standard model training can be defined as follows:
 {
\begin{equation}
L_{adv}(x,y)=\alpha_{nat}L(x,y)+\alpha_{adv}L(x+\delta,y),
\end{equation} 
}
where  {the $\alpha_{nat}$ and $\alpha_{adv}$ are the trade-off hyper parameters}, $\delta$ represents the adversarial perturbation with the maximum perturbation scale $\epsilon$, and $x$ and $y$ refer to the input samples and true labels, respectively. By performing a first-order Taylor expansion on the loss  $L_{adv}(x,y)$ for adversarial training, we can obtain:
 {
\begin{align}
\label{eq3-2-1}
    L_{adv}(x,y)&\approx\alpha_{nat}L(x,y)+\alpha_{adv}(L(x,y)+\delta \cdot \partial_xL(x,y)), \notag \\ 
    &=L(x,y)+\alpha_{adv}\delta \cdot \partial_xL(x,y),
\end{align}
}
where  $\cdot$ denotes the matrix point multiplication operation, $\partial_xL(x,y)$ represents the first-order partial derivative of $L(x,y)$ with respect to $x$. Following \cite{simon2019first}, further expansion of $\partial_xL(x,y)$ can be obtained as follows:
\begin{align}
   \delta \cdot \partial_xL(x,y)&=\mathop{\max}_{ \mid\mid\delta\mid\mid_p \leq \epsilon}  \mid L(x+\delta,y)- L(x,y) \mid, \notag \\ 
    &\approx\mathop{\max}_{ \mid\mid\delta  \mid\mid_p \leq \epsilon} \mid \delta \cdot \partial_xL(x,y) \mid, \notag \\
    &= \epsilon \mid\mid\partial_xL(x,y)\mid\mid_q,
\end{align}
 {where $|.|$ denotes the absolute value, the symbol $p$ or $q$ denotes the $l_p$ or $l_q$ norm, $\epsilon \mid\mid\partial_xL(x,y)\mid\mid_q$ is the dual norm of  $\mathop{\max}_{ \mid\mid\delta\mid\mid_p\leq\epsilon} |\delta\cdot\partial_xL(x,y)|$, while these two variables satisfy: $\frac{1}{q}+\frac{1}{p}=1$.}
% and $\mid\mid.\mid\mid_q$ can be defined as follows:
% \begin{equation}
% \epsilon \mid\mid\partial_xL(x,y)\mid\mid_q=sup \left\{\delta \mid\mid\partial_xL(x,y)\mid\mid_q \big| \ {\mid\mid \delta\mid\mid}_p\le\epsilon \right\},
% \end{equation} 
 { $\epsilon\mid\mid\partial_xL(x,y)\mid\mid_q$ represents the  regularization on the partial derivative of the optimization loss $L(x,y)$ with respect to the input data $x$.}  

%  {Actually, the regular term ${\mid\mid\partial_xL(x,y)\mid\mid}_q$ can be used to measure the sensitivity of the input sample $x$ in the model distribution space: when the value of ${\mid\mid\partial_xL(x,y)\mid\mid}_q$ is large, small perturbation in $x$ will lead to obvious fluctuation in the prediction results of the model; When the value of ${\mid\mid\partial_xL(x,y)\mid\mid}_q$ is small, the prediction results of the model are not sensitive to small perturbation in $x$. }

Based on the above analysis, the difference between the two training methods is that the optimization loss $L_{adv}(x,y)$ of the adversarial training adds a gradient regularization constraint $\epsilon\mid\mid\partial_xL(x,y)\mid\mid_q$ based on the optimization loss $L(x,y)$  of the standard training, and the gradient regularization term constraint will directly affect the update of the model weight. 
Specifically, in the training process, standard model weight $w_{nat}$ and robust model weights $w_{adv}$ is updated according to the gradient descent criterion:
\begin{equation}
\label{gradient1}
w_{nat}=w_{nat}-\eta \partial_{w_{nat}}L(x,y),
\end{equation}
\begin{equation}
\label{gradient2}
w_{adv}=w_{adv}-\eta \partial_{w_{adv}}L_{adv}(x,y).
\end{equation}
where $\eta$ is the learning rate of the weight, $\partial_{w_{nat}}L(x,y)$ and $\partial_{w_{adv}}L(x,y)$ is the partial derivative of the $L(x,y)$ relative to the weights $w_{nat}$ and $w_{adv}$. 
 {Then we assume that both the standard model and robust model have a weight $w_0$ randomly initialized with the same means and variances}. Based on Eq.(\ref{gradient1}) and Eq.(\ref{gradient2}), the expectations of standard model weight $w_{nat}$ and robust model weights $w_{adv}$ can be formulated as follows:
 {
\begin{equation}
E(w_{nat})=E(w_{0})-\eta E(\partial_{w_{0}}L(x,y)),
\end{equation}
\begin{equation}
E(w_{adv})=E(w_{0})-\eta E(\partial_{w_{0}}L_{adv}(x,y)),
\end{equation}
}
$\partial_{w_{0}}L(x,y)$ and $\partial_{w_{0}}L_{adv}(x,y)$ are the partial derivative of loss $L(x,y)$ and $L_{adv}(x,y)$ relative to the weights $w_{0}$. 
Then the weight distribution difference between the standard-trained model and the robust-trained model can be derived:
\begin{align}
E(\Delta_w)&=\eta \mid E(w_{adv})-E(w_{nat}) \mid, \notag \\
&=\eta \mid E(\partial_{w_0}L_{adv}(x,y)-\partial_{w_0}L(x,y)) \mid, \notag \\
&=\alpha_{adv}\epsilon\eta  E(\partial_{w_0}\mid\mid\partial_xL(x,y)\mid\mid_q),
\end{align}
specifically, when the perturbation scale $\epsilon$ or the adversarial trade-off hyper-parameter $\alpha_{adv}$ is 0, the model weight expectations obtained by adversarial training and standard training are consistent.  {If some special constraints are added to the adversarial perturbation scale $\epsilon$ and the adversarial examples are located on the distribution of clean examples, for example, the on-manifold adversarial examples \cite{stutz2019disentangling}, the model can improve the generalization of the model by fitting this difference term. However, when simultaneously considering clean examples and the adversarial examples far from the distribution of clean examples,  inherent weight distribution differences indeed exist between the standard-trained model and the robust-trained model. }

 {For robust-trained models, the model weight $w_{adv}$ is optimized with the regularization term, and the final prediction does not change significantly when $x$ is perturbed within perturbation scale $\epsilon$, in other words, the expectation difference $E(\Delta_w)$ for $w_{adv}$ can enhance the model robustness. However, after adding the additional regularization, the sensitivity of weight $w_{adv}$ to clean examples will also decrease, which leads to poor accuracy for clean examples.}

 {On the contrary, for the standard-trained model, the model weight $w_{nat}$ is only optimized by the loss $L(x,y)$ but without the additional regularization term. After the gradient descent operation, the model weight $w_{nat}$ does not take into account the change of the model prediction results when $x$ is slightly perturbed, which leads to weak robustness toward adversarial examples. So the expectation difference of weight distribution  $E(\Delta_w)$  caused by the optimization object directly leads to the trade-off problem.}

Based on the above observation and the theoretical analysis, we argue that it is difficult to contain two groups of weights with different distributions in a static network, and the weights’ difference is the necessary condition for a better trade-off. Inspired by dynamic networks, we consider to replace static networks with dynamic networks. A sample-wise dynamic network model can change the model weight according to the input samples, which provides the potential to solve the trade-off between accuracy and robustness.

\section{Adversarial Weight-Varied Network}
\label{AGnet}
\subsection{Overview}
The pipeline of the proposed framework is shown in Figure \ref{fig:framework}. We propose a sample-wise dynamic Adversarial Weight-Varied Network (AW-Net) to enhance the accuracy and robustness. AW-Net contains two branches: a dynamic weight sub-network and an adversarial router sub-network. The dynamic weight sub-network applies different filters' weights to adapt various inputs. Meanwhile, MixBN is used to handle the different distributions of clean and adversarial examples, respectively. The adversarial router sub-network is used to generate regulation signals to adjust the filters' weights of the dynamic weight sub-network.  {The theoretical analysis for AW-Net can be found in the Appendix.}

\subsection{Dynamic Weight Sub-network}
\label{Dynamic Weight Sub-network}

The analysis in Section \ref{differences} inspires us to design a dynamic network composed of multiple blocks that can dynamically adjust the filter weights. For the $j$-th block, we assume  $\psi_{j} \in \mathbb{R}^{c^{in}_j \times w_{j}^{in} \times h_{j}^{in}}$ is the input feature with $c^{in}_j$ channels, and $\phi_{j} \in \mathbb{R}^{c^{out}_j \times w_{j}^{out} \times h_{j}^{out}}$ is the output feature with $c^{out}_j$ channels. We utilize a residual structure to construct each block in AW-Net. The $j$-th block is defined as follows:
\begin{align}
\phi_{j} =\sigma(\mathcal{F}_{j}(\psi_{j})+\psi_{j}),
\end{align}
where $\sigma(\cdot)$ denotes the Relu layer. The symbol $\mathcal{F}_{j}(\cdot)$ contains the convolution layer $\mathcal{W}^{conv}_j \in \mathbb{R}^{c^{in}_j \times k_{j} \times k_{j} \times c^{out}_j}$ with kernel size $k_{j}$ and other nonlinear layers such as Batch-Normalization layer. $\mathcal{W}^{conv}_j$ consists of multiple filters $\mathcal{K}_m \in \mathbb{R}^{c^{in}_j \times k_{j} \times k_{j}}, m = (1,2,...,c^{out}_j)$, thus:
\begin{align}
\mathcal{W}^{conv}_j =concat(\mathcal{K}_1, ..., \mathcal{K}_{c_{out}}),
\end{align}
where $concat(\cdot)$ is the concatenate operation along with the fourth dimension $c^{out}_j$. Without losing generality, different filters in convolution layer $\mathcal{W}^{conv}_j$ can be viewed as relatively independent units for extracting feature information and are similar to the individual ``experts''.  
From this point of view, different filters $\mathcal{K}_m$ in our AW-Net block are supposed to have different importance when processing adversarial and clean samples. In order to make each filter play its specialty, we apply a filter-wise vector $\omega_{j} =(\omega^1_{j},...,\omega^{c_{out}}_j)$ to give them different regulation signals as follows: 
\begin{equation}
\mathcal{\hat{W}}^{conv}_j =concat(\omega^1_{j}\cdot\mathcal{K}_1, ..., \omega^{c_{out}}_j\cdot\mathcal{K}_{c_{out}}),
\end{equation}
then we perform a convolution operation $*$ to obtain the output feature:
\begin{equation}
\label{eq:2}
\pi_j = \mathcal{\hat{W}}^{conv}_j*\psi_{j}.
\end{equation} 

With the assistance of weight adjustment, the block can deal with adversarial and clean examples in a distinctive way of feature processing. For the adversarial example, the filters' weights will have small variances, and filters' weights will have large variances for clean examples, as the phenomenon is shown in Figure \ref{distribution}.

Batch Normalization \cite{ioffe2015batch} is an effective method to accelerate the training process and is usually attached after the convolution layer.  However, the existence of the BN layer will eliminate the feature difference between adversarial and clean examples, and make their distributions tend to be the same. The weight adjustment mechanism toward filters will not take advantage of handling different characteristics and may negatively influence the performance of the network, leading to the catastrophic results that both distributions cannot be fitted well.

Inspired by the gated BN~\cite{liu2020towards}, we design a  MixBN, where a Clean BN $BN_{nat}$ is used to handle the features $\pi_j^{nat}$ of clean examples $x_{nat}$ and Adv BN $BN_{adv}$ is used to handle the features $\pi_j^{adv}$ of adversarial examples $x_{adv}$. 
% The $BN_{adv}$ and $BN_{nat}$ can be formulated as follows:  
% \begin{equation}
%      \begin{split}
%      BN_{adv} = \gamma_{adv}\frac{ \psi_{conv}- \mu_{adv} }{\sqrt{\sigma_{adv}^2 + \epsilon}} + \beta_{adv}, &\psi_{conv} \in x_{adv} \\
%      BN_{nat} = \gamma_{nat}\frac{ \psi_{conv}- \mu_{nat} }{\sqrt{\sigma_{nat}^2 + \epsilon}} + \beta_{nat}, &\psi_{conv} \in x_{nat}
%      \end{split}
% \end{equation}
% where $ \gamma_{adv}$, $\beta_{adv}$, $ \gamma_{nat}$, $\beta_{nat}$ denote the learnable parameters of BN layers. $\mu_{adv}$, $\sigma_{adv}$ and $\mu_{nat}$, $\sigma_{nat}$ represent the means and variances of adversarial and clean features, respectively. $\epsilon$ denotes the minor deviation to prevent the denominator towards zero. 

In the training process of the MixBN structure, the category of features can be obtained and we perform an operation of type selection and separately train the Adv BN and Clean BN with adversarial examples $x_{adv}$ and clean examples $x_{nat}$. The training process of MixBN can be denoted as follows:
\begin{align}
\begin{split}
BN_{mix}= \left \{
\begin{array}{ll}
    BN_{adv}(\pi_j^{adv}),             &\pi^{adv}_j \propto x_{adv},\\
    BN_{nat}(\pi_j^{nat}),             &\pi_j^{nat} \propto x_{nat}.
\end{array}
\right.
\end{split}
\end{align}

In the testing process of the MixBN, we use type predictions $P_{type}$ (defined in the next section) from the adversarial router sub-network as different BN's weights and use both Adv BN and Clean BN to deal with input features. The testing process of MixBN can be denoted as follows:
% \begin{align}
% \begin{split}
%     BN_{mix} =(1-P_{type}) & BN_{adv}(\pi_j^{adv})  
%     \\&+ P_{type}BN_{nat}(\pi_j^{nat}).
% \end{split}
% \end{align}
\begin{align}
    BN_{mix} =(1-P_{type})  BN_{adv}(\pi_j^{adv})  
    + P_{type}BN_{nat}(\pi_j^{nat}).
\end{align}

After the above operations, $\mathcal{F}_{j}(\cdot)$ is defined as:
\begin{align}
\mathcal{F}_{j}(\psi_{j})=BN_{mix}(\mathcal{\hat{W}}^{conv}_j*\psi_{j}).
\end{align}

Here, we use a two-layer residual structure, thus the $j$-th block is finally formulated as follows:
\begin{align}
\phi_{j} =\sigma(\mathcal{F}_{j}(\sigma(\mathcal{F}_{j}(\psi_{j})))+\psi_{j}).
\end{align}

Here we replace the whole two-layer basic block with our AW-Net block mentioned based on the basic ResNet structure \cite{he2016deep}, and the dimension of the final fully-connected layer adapts to the $x$'s dimension of different applied dataset (e.g., CIFAR) by average pooling operation.

The prerequisite for an effective dynamic weight sub-network is based on the excellent performance of regulation signals $\omega_{j}$ and type predictions $P_{type}$ in the testing process. So how to design an effective adversarial router sub-network needs to be solved. 

\subsection{Adversarial Router Sub-network}
\label{Adversarial Detector Sub-network}

We design a lightweight sample-wise adversarial router sub-network, including a feature extractor and an adversarial weight regulator structure. The feature extractor is designed for getting semantic features from inputs and the following adversarial regulator structure is used to analyse the semantic features and generate guiding information toward dynamic weight sub-network.

 Here a backbone network is used as the feature extractor $\mathcal{D}$ to get the representative feature from origin images $x$. To facilitate subsequent processing for the weight regulator, we get a one-dimension feature map $\psi_{M} \in \mathbb{R}^{d}$ after feature extractor $\mathcal{D}$. 
 The operation can be denoted as follows:
\begin{equation}
\psi_{M} = \mathcal{D}(x).
\end{equation}

Then the feature map $\psi_{M}$ is processed by the adversarial weight regulator, which has a type head to get the type predictions $P_{type}$ of input samples and signal heads to generate regulation signals $\omega_{j}$. 

For the design of the type head structure, we use a fully connected layer $\mathcal{W}^{type} \in \mathbb{R}^{ 2\times d}$, and a softmax operation is applied to get the type predictions $P_{type}$. The type head structure can be formulated as follows:
\begin{align}
P_{type} = softmax(\mathcal{W}^{type} \psi_{M}).
\end{align}

For the design of the signal head structure, to make full use of the entire information, individual signal heads are applied to generate regulation signals $\omega_{j}$ for each $j$-th block. Thus it is called a multi-head adversarial router sub-network. It can adjust the filter weights following Eq.(\ref{eq:2}) and allow every block to extract different semantic information of adversarial and clean examples at the filter level. Due to the instability of the neural network, directly using the feature map as regulation signals may lead to huge fluctuations whether in the training or testing process. Hence, a normalization operation toward the signal head is supposed to reduce the volatility. Here we use the $ tanh$-based function to make regulation signals more stable as follows:
\begin{align}
\omega_{j} = \big[tanh(\mathcal{W}_{j}\psi_{M}) + 1\big]^\beta,
\label{eq:eq9}
\end{align}
where $\mathcal{W}_{j} \in \mathbb{R}^{ c_{out} \times d}$ denotes the fully connected layer for the  $j$-th block. $\beta$ is a trade-off parameter to control the strength for guidance, if $\beta$ is too high, the effect of weight adjustment will be too strong, and the network inference will be unstable. If $\beta$ is too low, it will be difficult to distinguish adversarial and clean examples. 

For the adversarial router sub-network, a one-dimensional vector $\psi_{M}$ is obtained after the process of a feature extractor $\mathcal{D}$,  {and the feature is composed of multiple intermediate layers.} Then the feature vector $\psi_{M}$ is processed by the type head and the signal head. 
After the process of type head $\mathcal{W}^{type}$, the feature vector $\psi_{M}$ transforms into a 2-dimensional vector to predict the type predictions $P_{type}$.
For the signal head $\mathcal{W}_{j}$ in $j$-th block, due to the two-layer block design in our setting, two individual signal heads $\mathcal{W}_{j}$ are applied for two convolution layers in a block $\mathcal{W}^{conv}_j$ respectively and other setting is following the origin setting. 

All in all, with the favorable support of the adversarial router sub-network, the dynamic weight sub-network can give full play to its ability to deal with adversarial and clean examples with different weights.

\subsection{Joint Adversarial Training}

Similar to the static networks, AW-Net can be trained using the existing adversarial training algorithms, such as PGD-AT \cite{madry2017towards}, TRADES \cite{zhang2019theoretically},  and so on. However, to make full use of the dynamic characteristic in AW-Net,  
%The inner maximization is to generate adversarial examples as follows: 
 we choose Multi-teacher adversarial distillation (MTARD) \cite{Zhao2022Enhanced} as the adversarial training algorithm to perform a joint training strategy. MTARD utilizes a clean teacher and a robust teacher to jointly train the network to achieve the balance between accuracy and robustness, which is highly compatible with our AW-Net. Specifically, we can apply these two teachers to train the different weights for the clean examples and adversarial examples, respectively.  Clean examples are trained by the strong standard teacher $T_{nat}$, and adversarial examples are trained by a strong adversarial teacher $T_{adv}$, which is formulated as follow: 
\begin{align}
\mathcal{L}_{adv} = \tau^2\mathcal{L}_{KL}(S^{\tau}(x_{adv}),T_{adv}^{\tau}(x_{adv})),\\ \mathcal{L}_{nat} = \tau^2\mathcal{L}_{KL}(S^{\tau}(x_{nat}),T_{nat}^{\tau}(x_{nat})),
\end{align}
where $\mathcal{L}_{nat}$ and $\mathcal{L}_{adv}$ denote adversarial distillation loss for clean and adversarial teacher, respectively, $\mathcal{L}_{KL}$ denotes the Kullback-Leibler Divergence loss, $S(\cdot)$ denotes AW-Net, and $S^{\tau}(\cdot)$ denotes the tempered variant of $S(\cdot)$ with temperature $\tau$ \cite{hinton2015distilling}. Also, we design a classification loss to train the adversarial router sub-network to get type predictions. The minimization loss of training the AW-Net is as follows:
\begin{align}
\label{eq12}
\mathcal{L}_{total} = \mathcal{L}_{type}(P_{type},y') + \alpha_1 \mathcal{L}_{adv}+ \alpha_2 \mathcal{L}_{nat},
\end{align}
where $\mathcal{L}_{type}$ denotes the type loss for the binary classification,  {here we apply binary Cross-Entropy loss}.  $y'$ is a label that can represent adversarial and clean types. $\alpha_{1}$ and $\alpha_{2}$ are the hy-parameters, 
% to balance the trade-off as mentioned in the original paper  \cite{Zhao2022Enhanced}.
 which will be automatically adjusted in the training process as described in their original paper  \cite{Zhao2022Enhanced}. Therefore, we don't manually tune them in the experiment.

% {It should be mentioned that previous state-of-the-art adversarial detection methods \cite{cohen2020detecting,abusnaina2021adversarial} require training different types of detectors for adversarial examples generated by different attack methods, which leads to poor generalization. When the detector is trained by adversarial examples generated by a specific attack, its performance will be greatly reduced when facing other types of attack methods. Out of consideration to maximize the generalization, we need to generate some diverse adversarial examples to train our AW-Net to improve the generalization of adversarial router sub-network. If adversarial detection mechanism with better generalization exists in the future, this training strategy can also be adjusted and optimized.}

\begin{table*}[ht]
\caption{White-box robustness on CIFAR-10  datasets. All the models are trained following one of the state-of-the-art adversarial training method MTARD \cite{Zhao2022Enhanced}. All the results are the best checkpoints based on W-Robust Acc.}
\label{table:white-cifar10}
\scalebox{0.95}  { 
\begin{tabular}{
>{\columncolor[HTML]{FFFFFF}}c 
>{\columncolor[HTML]{FFFFFF}}c |
>{\columncolor[HTML]{FFFFFF}}c |
>{\columncolor[HTML]{FFFFFF}}c 
>{\columncolor[HTML]{FFFFFF}}c |
>{\columncolor[HTML]{FFFFFF}}c 
>{\columncolor[HTML]{FFFFFF}}c |
>{\columncolor[HTML]{FFFFFF}}c 
>{\columncolor[HTML]{FFFFFF}}c |
>{\columncolor[HTML]{FFFFFF}}c 
>{\columncolor[HTML]{FFFFFF}}c |
>{\columncolor[HTML]{FFFFFF}}c 
>{\columncolor[HTML]{FFFFFF}}c }
\hline 
% \noalign{\smallskip}
&            & Clean               & \multicolumn{2}{c|}{\cellcolor[HTML]{FFFFFF}FGSM} & \multicolumn{2}{c|}{\cellcolor[HTML]{FFFFFF}PGD$_{\scriptscriptstyle \rm{sat}}$} & \multicolumn{2}{c|}{\cellcolor[HTML]{FFFFFF}PGD$_{\scriptscriptstyle \rm{trades}}$} & \multicolumn{2}{c|}{\cellcolor[HTML]{FFFFFF}CW$_{\scriptscriptstyle \rm{\infty}}$} & \multicolumn{2}{c}{\cellcolor[HTML]{FFFFFF}AA} \\ 
% \noalign{\smallskip}
\hline 
% \noalign{\smallskip}
Model                                                    & Params & $\mathcal{A}_{nat}$ & $\mathcal{A}_{adv}$      & $\mathcal{A}_{w}$      & $\mathcal{A}_{adv}$        & $\mathcal{A}_{w}$        & $\mathcal{A}_{adv}$          & $\mathcal{A}_{w}$         & $\mathcal{A}_{adv}$     & $\mathcal{A}_{w}$     & $\mathcal{A}_{adv}$     & $\mathcal{A}_{w}$    \\  
% \noalign{\smallskip} 
\hline \noalign{\smallskip}
ResNet-34\cite{he2016deep}              & 20.3M    & 90.30\%             & 58.18\%                  & 74.24\%                & 45.56\%                    & 67.93\%                  & 48.95\%                      & 69.63\%                   & 45.95\%                 & 68.13\%               & 43.39\%               & 66.85\%            \\ 
ResNet-50\cite{he2016deep}              & 22.6M    & 90.22\%             & 60.65\%                  & 75.44\%                & 47.26\%                    & 68.74\%                  & 51.27\%                      & 70.75\%                   & 47.44\%                 & 68.83\%               &   44.70\%                      &  67.46\%                    \\
WideResNet-34-8\cite{zagoruyko2016wide} & 29.5M    & 90.14\%             & 62.23\%                  & 76.19\%                & 50.74\%                    & 70.44\%                  & 53.69\%                      & 71.92\%                   & 50.06\%                 & 70.10\%               & 46.63\%               &  68.39\%            \\
VGG-16-BN\cite{simonyan2014very}        & 32.2M    & 86.89\%             & 57.49\%                  & 72.19\%                & 44.53\%                    & 65.71\%                  & 48.46\%                      & 67.68\%                   & 44.22\%                 & 65.56\%               &    41.17\%                     &   64.03\%                   \\
RepVGG-A2\cite{ding2021repvgg}          & 25.6M    & 81.05\%             & 47.35\%                  & 64.20\%                & 35.17\%                    & 58.11\%                  & 38.67\%                      & 59.86\%                   & 34.74\%                 & 67.90\%               & 31.23\%               & 56.14\%            \\
\textbf{AW-Net (ours)}                  & 26.3M    & 93.08\%             & 77.38\%                  & \textbf{85.23\%}                & 50.93\%                    & \textbf{72.01\%}                  & 52.37\%                      & \textbf{72.73\%}                   & 47.82\%                 & \textbf{70.45\%}               &  44.56\%                       &   \textbf{68.82\%}                   \\ \noalign{\smallskip} \hline
\end{tabular}
}
\end{table*}

\begin{table*}[ht]
\caption{White-box robustness on  CIFAR-100 datasets. All the models are trained following one of the state-of-the-art adversarial training method MTARD \cite{Zhao2022Enhanced}. All the results are the best checkpoints based on W-Robust Acc.}
\label{table:white-cifar100}
\scalebox{0.95}  { 
\begin{tabular}{
>{\columncolor[HTML]{FFFFFF}}c 
>{\columncolor[HTML]{FFFFFF}}c |
>{\columncolor[HTML]{FFFFFF}}c |
>{\columncolor[HTML]{FFFFFF}}c 
>{\columncolor[HTML]{FFFFFF}}c |
>{\columncolor[HTML]{FFFFFF}}c 
>{\columncolor[HTML]{FFFFFF}}c |
>{\columncolor[HTML]{FFFFFF}}c 
>{\columncolor[HTML]{FFFFFF}}c |
>{\columncolor[HTML]{FFFFFF}}c 
>{\columncolor[HTML]{FFFFFF}}c |
>{\columncolor[HTML]{FFFFFF}}c 
>{\columncolor[HTML]{FFFFFF}}c }
\hline 
% \noalign{\smallskip}
&            & Clean               & \multicolumn{2}{c|}{\cellcolor[HTML]{FFFFFF}FGSM} & \multicolumn{2}{c|}{\cellcolor[HTML]{FFFFFF}PGD$_{\scriptscriptstyle \rm{sat}}$} & \multicolumn{2}{c|}{\cellcolor[HTML]{FFFFFF}PGD$_{\scriptscriptstyle \rm{trades}}$} & \multicolumn{2}{c|}{\cellcolor[HTML]{FFFFFF}CW$_{\scriptscriptstyle \rm{\infty}}$} & \multicolumn{2}{c}{\cellcolor[HTML]{FFFFFF}AA} \\ 
% \noalign{\smallskip} 
\hline 
% \noalign{\smallskip}
Model                                                    & Params & $\mathcal{A}_{nat}$ & $\mathcal{A}_{adv}$      & $\mathcal{A}_{w}$      & $\mathcal{A}_{adv}$        & $\mathcal{A}_{w}$        & $\mathcal{A}_{adv}$          & $\mathcal{A}_{w}$         & $\mathcal{A}_{adv}$     & $\mathcal{A}_{w}$     & $\mathcal{A}_{adv}$     & $\mathcal{A}_{w}$    \\ 
% \noalign{\smallskip} 
\hline 
\noalign{\smallskip}
ResNet-34\cite{he2016deep}              & 20.3M    & 65.69\%             & 33.47\%                  & 49.58\%                & 25.85\%                    & 45.77\%                  & 27.77\%                      & 46.73\%                   & 24.67\%                 & 45.18\%               & 21.87\%               & 43.78\%            \\
ResNet-50\cite{he2016deep}              & 22.6M    & 69.59\%             & 33.68\%                  & 51.64\%                & 25.02\%                    & 47.31\%                  & 27.38\%                      & 48.49\%                   & 25.36\%                 & 47.48\%               &  20.03\%                       &  44.81\%                    \\
WideResNet-34-8\cite{zagoruyko2016wide} & 29.5M    & 60.46\%             & 25.67\%                  & 43.07\%                & 20.05\%                    & 40.26\%                  & 21.70\%                      & 41.08\%                   & 18.84\%                 & 39.65\%               & 14.41\%               & 37.44\%            \\
VGG-16-BN\cite{simonyan2014very}        & 32.2M    & 57.65\%             & 28.23\%                  & 42.94\%                & 20.88\%                    & 39.27\%                  & 22.83\%                      & 40.24\%                   & 20.02\%                 & 38.84\%               &  18.35\%                       &  38.00\%                    \\
RepVGG-A2\cite{ding2021repvgg}          & 25.6M    & 51.44\%             & 24.23\%                  & 37.84\%                & 17.90\%                    & 34.67\%                  & 19.33\%                      & 35.39\%                   & 16.79\%                 & 34.12\%               & 15.04\%               & 33.24\%           \\
\textbf{AW-Net (ours)}                  & 26.3M    & 73.98\%             & 52.07\%                  & \textbf{63.03\%}                & 26.47\%                    & \textbf{50.23\%}                  & 27.51\%                      & \textbf{50.75\%}                   & 23.92\%                 & \textbf{48.95\%}                &  17.48\%                       &   \textbf{45.73\%}                   \\ \noalign{\smallskip} \hline
\end{tabular}
}
\end{table*}

\section{Experiment}
\label{experiments}

\subsection{Experimental Settings}
\textbf{Datasets:}
We conduct the experiments on  {three} public datasets, including CIFAR-10 \cite{krizhevsky2009learning}, CIFAR-100 and Tiny-ImageNet \cite{le2015tiny}. For CIFAR-10 and CIFAR-100, the image size is 32 $\times$ 32, the training set contains 50,000 images, and the testing set contains 10,000 images. CIFAR-10 covers 10 classes while CIFAR-100 covers 100 classes. 
 {Tiny-ImageNet dataset is a subset of ImageNet dataset \cite{deng2009imagenet}, and the image size is resized from  224 $\times$ 224 (original size of ImageNet) to 64 $\times$ 64.} The training set contains 100,000 images, and the testing set contains 10,000 images, covering 200 classes.

\noindent \textbf{Compared SOTA methods}: Because our method proposes a novel network architecture, we compare it with other state-of-the-art network architectures with similar parameter scales. To ensure fair comparisons, all the networks are trained with the same optimization algorithm. Specifically, for CIFAR-10 and CIFAR-100, the compared networks include ResNet-34 \cite{he2016deep}, ResNet-50 \cite{he2016deep}, WideResNet-34-8 \cite{zagoruyko2016wide}, VGG-16-BN \cite{simonyan2014very} and RepVGG-A2 \cite{ding2021repvgg}. In addition, we also compare with the new transformer architecture: Swin Transformer \cite{liu2021swin} as well as other dynamic networks: RDI-ResNet-38 \cite{hu2020triple} and ADVMOE \cite{zhang2023robust}. 

% For Tiny-ImageNet dataset, we compare with the PreActResNet-34 \cite{he2016identity}. Our AW-Net and PreActResNet-34 \cite{he2016identity} are trained with another state-of-the-art optimization algorithm TRADES \cite{zhang2019theoretically}. 

\noindent \textbf{Adversarial Attack methods}: We evaluate AW-Net and other comparison models against four classic white-box attack methods: FGSM \cite{goodfellow2014explaining}, PGD$_{\scriptscriptstyle \rm{sat}}$ \cite{madry2017towards}, PGD$_{\scriptscriptstyle \rm{trades}}$ \cite{zhang2019theoretically}, CW$_{\scriptscriptstyle \rm{\infty}}$ \cite{carlini2017towards}. The total step is 20, the difference between the two PGD attacks lies in the step size (0.003 for PGD$_{\scriptscriptstyle \rm{trades}}$ 
 and 2/255 for PGD$_{\scriptscriptstyle \rm{sat}}$). The step of CW$_{\scriptscriptstyle \rm{\infty}}$ is 30, and the perturbation of PGD and CW$_{\scriptscriptstyle \rm{\infty}}$ is bounded to the $L_{\infty}$ norm $\epsilon$ = 8/255. {Meanwhilie, we also evaluate different methods against the strong AutoAttack (AA) \cite{croce2020reliable} method.}  {It should be mentioned that CW attack and two types of attacks in AutoAttack (including FAB-Attack and Square Attack) are unseen adversarial attacks for our AW-Net.}

\noindent \textbf{Evaluation metric:}
We use the Weighted Robust Accuracy (W-Robust Acc) following \cite{gurel2021knowledge} to comprehensively evaluate the accuracy and robustness, which can be formulated:
\begin{align}
\mathcal{A}_w = \gamma_{nat}\mathcal{A}_{nat} +\gamma_{adv}\mathcal{A}_{adv},
\end{align}
where $\mathcal{A}_w$ is the W-Roubst Acc computed from the clean accuracy $\mathcal{A}_{nat}$ and the adversarial accuracy $\mathcal{A}_{adv}$.  {Here we argue that the clean accuracy $\mathcal{A}_{nat}$ and adversarial robustness $\mathcal{A}_{adv}$ are equally important in actual application. Based on this expectation, the hyper-parameters $\gamma_{nat}$ and $\gamma_{adv}$ are set to 1/2, which can better reflect the trade-off between the accuracy and robustness.}

\noindent \textbf{Implementation details:}
On CIFAR-10 and CIFAR-100 datasets, we use ResNet-18 as the basis to construct the dynamic weight sub-network, and then modify each residual block according to Section \ref{Dynamic Weight Sub-network},  {which includes 8 residual blocks; for the adversarial router sub-network, we also use ResNet-18 as the feature extractor, which also includes 8 residual blocks.}  Note that users can also adapt to other networks like ResNet-34 and ResNet-50 as the basis. The training optimizer is Stochastic Gradient Descent (SGD) optimizer with momentum of 0.9 and weight decay of 2e-4, and the initial learning rate is 0.1 for the dynamic weight network and 0.01 for the adversarial router sub-network.

% we select the two first white-box attacks (targeted and untargeted APGD attacks) in \cite{croce2020reliable} to generate the diverse adversarial examples. Considering the computational cost, 

 {For the training process,  we train our AW-Net for 100 epochs, and the learning rate is divided by 10 at the  75-th, and 90-th epoch, and the batch size is 256.} The perturbation is bounded to the $L_{\infty}$ norm $\epsilon$ = 8/255.  $\beta$ is set to 3. And $\tau$ is set to 1 for CIFAR-10 and 5 for CIFAR-100. The teacher models on CIFAR-10 and CIFAR-100 follow the setting in \cite{Zhao2022Enhanced}. 
 %  {For Tiny-ImageNet, we train the above models for 50 epochs, and the learning rate is divided by 25-th and 40-th epochs. The batch size is 128 and $\tau$ is set to 5.  And the clean teacher is standard-trained ResNet-34, while the robust teacher model is ResNet-34 trained by \cite{zhang2019theoretically}.} 
 {For the baselines, we strictly follow the setting in \cite{Zhao2022Enhanced}. }

\subsection{Ablation Study} 

In order to verify the effectiveness of each component in our AW-Net, we conduct a set of ablation studies. The experiments are conducted on the CIFAR-10 dataset. 

\subsubsection{Effects of different components}
We explore three components to test their effects. (1) 
Initially, we apply the adversarial router sub-network to generate adversarial regulation signals and guide the dynamic weight network to distinguish clean and adversarial examples for different filters but without MixBN structure.  {More specifically, the adversarial router sub-network only consists of the signal head but does not consist of the type head}. (2) Then, we explore the improvement brought by the MixBN structure but without type predictions from the adversarial router sub-network in the testing process, which only uses the Adv BN.  {Actually, we can select the clean BN as another alternative, however, due to the poor performance when facing the adversarial examples, we only report the performance of using adv BN}. (3) Furthermore, we use type predictions as MixBN weights, which is the final structure of our AW-Net. We also report the performance of ResNet-18 as the baseline.  The results are shown in Table \ref{table:3}.

\setlength{\tabcolsep}{1pt}
\begin{table}[ht]
\begin{center}
\caption{Ablation study towards each component of AW-Net. ``Dynamic Weight'' denotes the dynamic weight sub-network only guided by signal head in adversarial router; ``MixBN'' represents using the MixBN layer in the training process but applying a single BN in the testing process; ``Type Head'' denotes using the type predictions in the adversarial router sub-network as the weights of MixBN in the testing process. ``Baseline" is the basis network structure to construct AW-Net.}
\label{table:3}
\begin{tabular}{m{1.4cm}<{\centering}|m{1.6cm}<{\centering}m{1.2cm}<{\centering}m{1.2cm}<{\centering}|m{1.2cm}<{\centering}}
\hline
 Attack & Dynamic Weight  &  MixBN  &  Type Head & $\mathcal{A}_{w}$  \\
\hline
\multirow{4}*{PGD$_{\scriptscriptstyle \rm{trades}}$} & \checkmark & $\times$ & $\times$  & 68.03\% \\
~ & \checkmark & \checkmark & $\times$  &  67.42\% \\
~ & \checkmark & \checkmark & \checkmark  &\textbf{72.73\%} \\
\cline{2-5}
~ &~&Baseline&~&70.48\% \\
\hline
\multirow{4}*{CW$_{\scriptscriptstyle \rm{\infty}}$} &  \checkmark & $\times$ & $\times$  & 66.05\%  \\
~ & \checkmark & \checkmark & $\times$ & 65.25\% \\
~ & \checkmark & \checkmark & \checkmark   &\textbf{70.45\%} \\
\cline{2-5}
~ &~&Baseline&~ &67.97\% \\
\hline
\end{tabular}
\end{center}
% \vspace{-0.5cm}
\end{table}
\setlength{\tabcolsep}{1.4pt}

From Table \ref{table:3}, we see that only using the dynamic weight cannot achieve good performance. After adding the MixBN in the training process but applying a single BN in the testing process, the result is still not particularly ideal. But when the MixBN is weighted with the type predictions in the testing process, the performance is obviously improved, showing the necessity of using MixBN to process features from different weight distributions, and it also indirectly shows the effectiveness of the adversarial router sub-network. As a comparison with the baseline ResNet-18  network, our dynamic network achieves a nearly 2.5\% improvement, showing the effectiveness of our idea.

\setlength{\tabcolsep}{1pt}
\begin{table}[t]
\begin{center}
\caption{The comparison of the separate combination (Separate-Net) and AW-Net. The Separate-Net consists of a standard-trained model and a robust-trained model, and the output is selected by a trained detector. }
\label{table:simple}
\scalebox{0.96}  { 
\begin{tabular}{
>{\columncolor[HTML]{FFFFFF}}c |
>{\columncolor[HTML]{FFFFFF}}c |
>{\columncolor[HTML]{FFFFFF}}c 
>{\columncolor[HTML]{FFFFFF}}c |
>{\columncolor[HTML]{FFFFFF}}c 
>{\columncolor[HTML]{FFFFFF}}c }
\hline 
% \noalign{\smallskip}
 &         Clean               & \multicolumn{2}{c|}{\cellcolor[HTML]{FFFFFF}PGD$_{\scriptscriptstyle \rm{trades}}$} & \multicolumn{2}{c}{\cellcolor[HTML]{FFFFFF}CW$_{\scriptscriptstyle \rm{\infty}}$} \\ 
 % \noalign{\smallskip} 
 \hline 
 % \noalign{\smallskip}
Model                                                    & $\mathcal{A}_{nat}$ & $\mathcal{A}_{adv}$      & $\mathcal{A}_{w}$     & $\mathcal{A}_{adv}$     & $\mathcal{A}_{w}$    \\ 
% \noalign{\smallskip} 
\hline \noalign{\smallskip}
Separate-Net    & 93.30\%             & 50.11\%                  & 71.71\%               & 45.35\%                 & 69.33\%              \\
Separate-Net-L         & 93.34\%             & 50.83\%                  & 72.09\%               & 45.97\%                 & 69.66\%              \\
\textbf{AW-Net(ours)}           & 93.08\%             & 52.37\%                  & 72.73\%               & 47.82\%                 & 70.45\%              \\
\textbf{AW-Net-L (ours)}          & 92.50\% &   55.23\% & \textbf{73.87\%} & 48.76\% & \textbf{70.63\%}          \\ \noalign{\smallskip} \hline
\end{tabular}
}
\end{center}
\end{table}
\setlength{\tabcolsep}{1.4pt}

% \setlength{\tabcolsep}{1pt}
% \begin{table}[ht]
% \begin{center}
% \caption{ {The comparison of the separate combination (Separate-Net) and AW-Net. The Separate-Net consists of a standard-trained model and a robust-trained model, and the output is selected by a trained detector. } }
% \label{table:simple}
% \begin{tabular}{m{1.6cm}<{\centering}m{2.8cm}<{\centering}|m{1.2cm}<{\centering}m{1.2cm}<{\centering}m{1.2cm}<{\centering}}
% \hline
%  Attack & Model &$\mathcal{A}_{nat}$ &$\mathcal{A}_{adv}$& $\mathcal{A}_{w}$  \\
% %\hline
% %\multirow{2}*{PGD$_{\scriptscriptstyle \rm{sat}}$}  & Simple Combination & 93.30\% & 48.35\% & 70.83\% \\
% %~ & \textbf{AW-Net(ours)} & 93.08\% & 50.93\% & \textbf{72.01\%} \\
% \hline
% \multirow{4}*{PGD$_{\scriptscriptstyle \rm{trades}}$}  & Separate-Net & 93.30\% & 50.11\% & 71.71\% \\
% ~ & Separate-Net-L & 93.34\% & 50.83\% & 72.09\% \\
% ~ & \textbf{AW-Net(ours)} & 93.08\% & 52.37\% & 72.73\% \\
% ~ & \textbf{AW-Net-L(ours)} & 92.50\% & 55.23\% & \textbf{73.87\%}  \\
% \hline
% \multirow{4}*{CW$_{\scriptscriptstyle \rm{\infty}}$}  & Separate-Net & 93.30\% & 45.35\% & 69.33\% \\
% ~ & Separate-Net-L & 93.34\% & 45.97\% & 69.66\% \\
% ~ & \textbf{AW-Net(ours)} & 93.08\% & 47.82\% & 70.45\%  \\
% ~ & \textbf{AW-Net-L(ours)} & 92.50\% & 48.76\% &  \textbf{70.63\%} \\
% \hline
% \end{tabular}
% \end{center} 
% \end{table}
% \setlength{\tabcolsep}{1.4pt}

\subsubsection{Effects of joint design for two branches}
\label{Effects of joint design}
 {To prove the superiority of our designed filter-level dynamic weight strategy for the two branches in our architecture, we compare AW-Net with a separate combination method (Separate-Net).  {Specifically, we replace the dynamic weight sub-network with} a standard-trained model (MobileNet-v2) and a robust-trained model (ResNet-18 trained by \cite{zhang2019theoretically}) for inference, and the output is determined by a trained and thresholded detector (the structure is the same with our adversarial router sub-network in AW-Net), and the model scale of this Separate-Net is similar to our AW-Net.  {Meanwhile, we also compare the large-version AW-Net-L and Separate-Net-L by replacing ResNet-18 with ResNet-34.} 
 
 The result in Table \ref{table:simple} shows the superiority of AW-Net compared with the Separate-Net.  Under the attack of PGD$_{\scriptscriptstyle \rm{trades}}$ and CW$_{\scriptscriptstyle \rm{\infty}}$, our AW-Net has 1.02\% and  1.12\% improvement compared to Separate-Net,  {while our AW-Net-L has 1.78\% and  0.97\% improvement compared to Separate-Net-L}. }
 {
Please note that the simple combination baseline is also designed by us based on our proposed ``divide and conquer'' weight strategy. The main difference between our AW-Net and this baseline is that AW-Net divides and conquers the weights at the filter level, while the simple combination baseline  divides and conquers at the entire network level. Because the core idea is similar, no significant difference exists between those two methods in the final result. This ablation study is used to verify our designed filter-level dynamic weight strategy in our AW-Net, and experiments show our AW-Net is indeed effective.
}

\subsubsection{Effects of different adversarial routers}

 {We also compare our adversarial router sub-network with current state-of-the-art adversarial detection mechanisms:  NNIF \cite{cohen2020detecting} and LNG \cite{abusnaina2021adversarial}. Here we apply the AUC score of the detection ROC curve as the metric to evaluate the performance of the adversarial detection method. The result in Table \ref{detector} shows our adversarial router sub-network can achieve nearly 100\% AUC score to distinguish clean examples and adversarial examples, even when the adversarial examples are generated by different attack methods. Specifically, under the attacks of PGD$_{\scriptscriptstyle \rm{trades}}$, our method has a 1.62\% improvement compared to the best baseline (NNIF). These results verify the advantages of our method compared with other state-of-the-art methods and show good generalization ability to different input examples.} 
\begin{table}[ht]
\begin{center}
\caption{ {Comparison of AUC scores (\%) for various adversarial detection methods and our adversarial router sub-network.}}
\label{detector}
\begin{tabular}{m{2.8cm}<{\centering}|m{1.1cm}<{\centering}m{1.1cm}<{\centering}m{1.1cm}<{\centering}}
\hline
 Detector  & FGSM & PGD$_{\scriptscriptstyle \rm{trades}}$ & CW$_{\scriptscriptstyle \rm{\infty}}$  \\
\hline
 NNIF \cite{cohen2020detecting}  & 99.96\% & 98.31\% & 99.50\% \\
 LNG \cite{abusnaina2021adversarial}  & 99.88\% & 91.39\% & 89.74\% \\
\textbf{AW-Net (ours)} & \textbf{99.99\%} &\textbf{ 99.93\%} & \textbf{99.89\%}  \\
\hline
\end{tabular}
\end{center}
% \vspace{-0.5cm}
\end{table}

\subsubsection{Effects of different $\beta$ values}

\begin{figure}[ht]
  \centering
\includegraphics[width=0.28\textwidth]{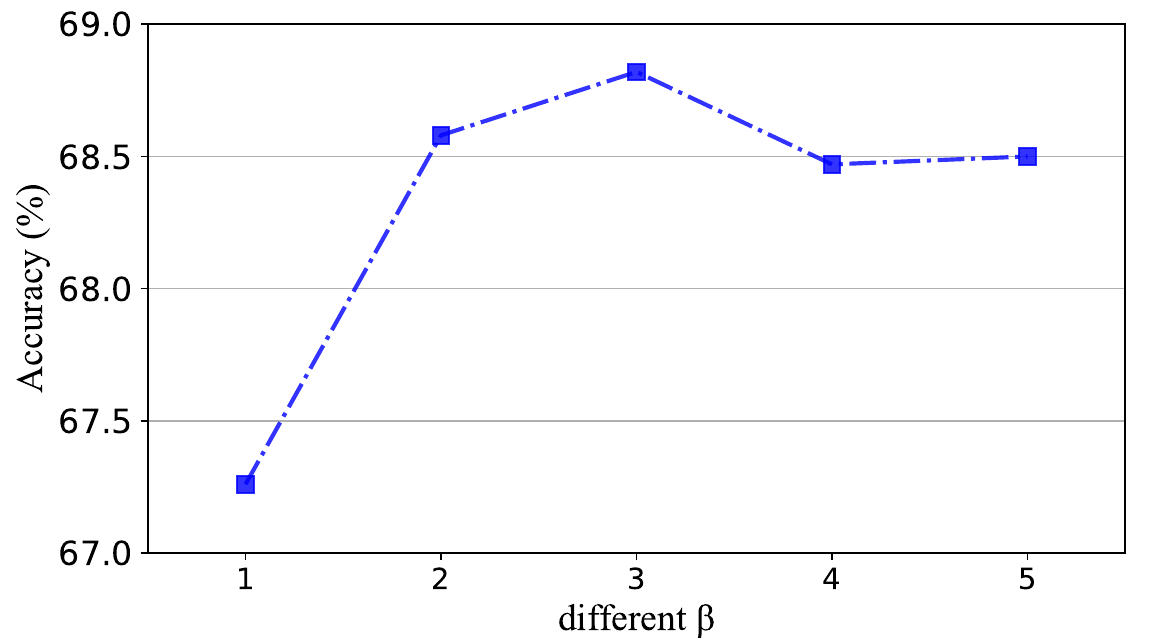}\\
\caption{ {The performance of AW-Net with different $\beta$. All the results are the best checkpoints based on $\mathcal{A}_{w}$. $\mathcal{A}_{w}$ is evaluated by AA.}}
\label{different beta}
\end{figure}

Here we explore the role of $\beta$ in  {Eq. (\ref{eq:eq9})} used in our AW-Net. We select the AW-Net trained on CIFAR-10 and test the results using different $\beta$ values. The results are shown in Figure \ref{different beta}. The results show that the proper parameter settings have an influence on the final results. If $\beta$ is too high, the network will fluctuate and lead to a lower performance. If $\beta$ is too small, the adversarial and clean examples will be not recognized well.  Finally, we choose the $\beta$ as 3 in our setting for CIFAR-10 and CIFAR-100.

% \setlength{\tabcolsep}{1pt}
% \begin{table}[ht]
% \begin{center}
% \caption{The performance of AW-Net with different $\beta$. All the results are the best checkpoints based on $\mathcal{A}_{w}$.}
% \label{table:6}
% \begin{tabular}{m{1.2cm}<{\centering}m{2cm}<{\centering}|m{1.2cm}<{\centering}m{1.2cm}<{\centering}m{1.4cm}<{\centering}}
% \hline
%  Attack & $\beta$ value &$\mathcal{A}_{nat}$ &$\mathcal{A}_{adv}$& $\mathcal{A}_{w}$  \\
% \hline
% \multirow{5}*{AA}  & $\beta = 1$ & 90.63\% & 43.89\% & 67.26\% \\
% ~ & $\beta = 2$ & 93.00\% & 44.16\% & 68.58\% \\
% ~ & $\beta = 3$ & 93.08\% &44.56\% &\textbf{68.82\%}  \\
% ~ & $\beta = 4$ & 93.39\% &43.54\% & 68.47\%  \\
% ~ & $\beta = 5$ & 93.09\% & 43.91\% & 68.50\%  \\
% \hline
% \end{tabular}
% \end{center} 
% \end{table}
% \setlength{\tabcolsep}{1.4pt}

\subsubsection{Effects of different training methods}

\begin{figure}[t]
  \centering
\includegraphics[width=0.33\textwidth]{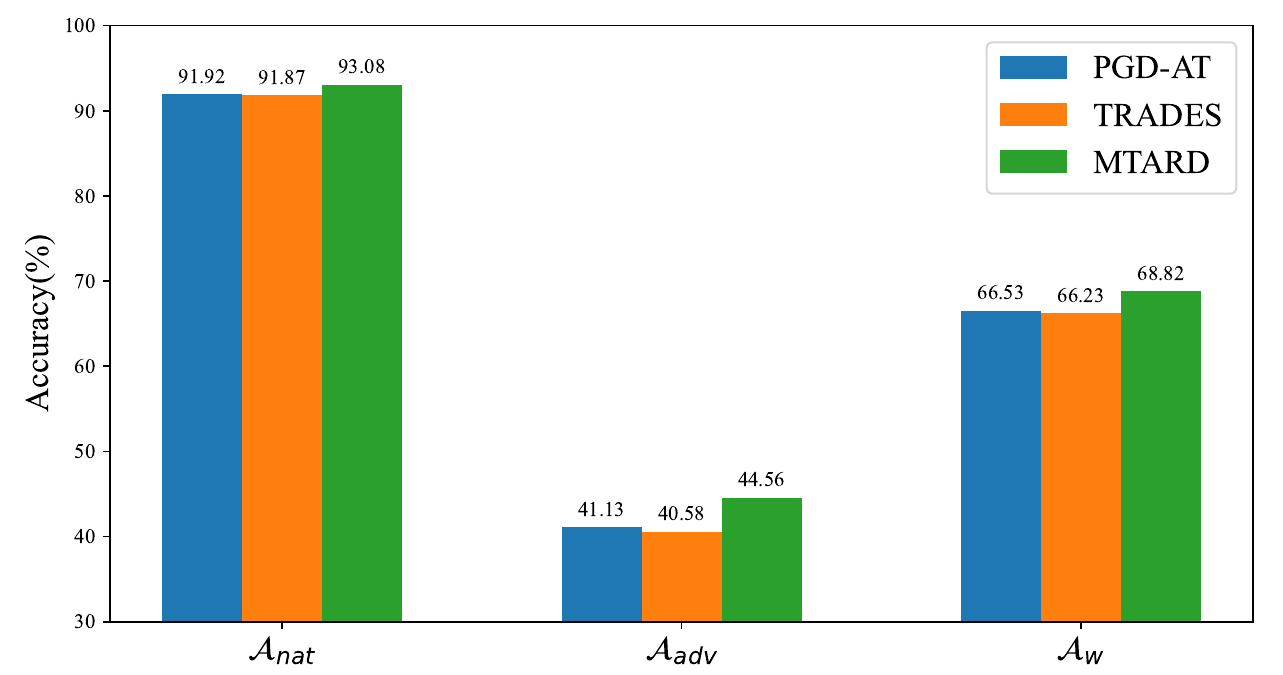}\\
\caption{{The performance of AW-Net with different training method. All the results are the best checkpoints based on $\mathcal{A}_{w}$. $\mathcal{A}_{w}$ is evaluated by AA.}}
\label{table:different training baseline}
\end{figure}

Here we explore the effect of different training algorithms on our AW-Net. We choose three different state-of-the-art methods, PGD-AT \cite{madry2017towards}, TRADES \cite{zhang2019theoretically}, and MTARD \cite{Zhao2022Enhanced}, and shows the performance on CIFAR-10. The results in Figure \ref{table:different training baseline}  show the model based on MTARD has better performances than TRADES and PGD-AT in both accuracy and robustness. In the metric of $\mathcal{A}_{w}$, MTARD can bring 2.29\% improvement under the evaluation of AA. This verifies that MTARD is suitable to train our dynamic network.

% \setlength{\tabcolsep}{1pt}
% \begin{table}[ht]
% \begin{center}
% \caption{ {The performance of AW-Net with different training method. All the results are the best checkpoints based on $\mathcal{A}_{w}$.}}
% \label{table:different training baseline}
% \begin{tabular}{m{1.2cm}<{\centering}m{2.4cm}<{\centering}|m{1.2cm}<{\centering}m{1.2cm}<{\centering}m{1.2cm}<{\centering}}
% \hline
%  Attack & Training Method &$\mathcal{A}_{nat}$ &$\mathcal{A}_{adv}$& $\mathcal{A}_{w}$  \\
% \hline
% \multirow{3}*{AA}  & PGD-AT \cite{madry2017towards} & 91.92\% & 41.13\% & 66.53\% \\
% ~ & TRADES \cite{zhang2019theoretically} & 91.87\% & 40.58\% & 66.23\% \\
% ~ & MTARD \cite{Zhao2022Enhanced} & 93.08\% &44.56\% &\textbf{68.82\%}  \\
% \hline
% \end{tabular}
% \end{center} 
% \end{table}
% \setlength{\tabcolsep}{1.4pt}

 {Here we provide an explanation of why MTARD achieves better performance for our AW-Net. The previous methods \cite{zhang2019theoretically,madry2017towards} often focus on a static weight network, and their ideas can often be understood as pursuing a balance between robust weight distribution and clean weight distribution. This type of idea is not to divide and conquer different weight distributions, but to treat them as a uniform but compromised weight distribution, which is not in line with our original intention of designing a dynamic weight network. MTARD attempts to use a clean teacher and a robust teacher as guidance for the two weight distributions, and this divide-and-conquer idea is highly consistent with our method, which is one of the reasons why we chose this method as our training baseline. We do not rule out that better optimization methods based on the idea of divide-and-conquer exist to train our AW-Net in the future, and this is also one of the directions for our further exploration.}

\subsection{Robustness on CIFAR-10 and CIFAR-100}
In this section, we comprehensively test the adversarial robustness of different network architectures against state-of-the-art white-box attacks and black-box attacks, respectively.

\subsubsection{Comparisons with static networks}

\textbf{Robustness against White-box Attacks}. 
Based on the results in Table \ref{table:white-cifar10} and Table \ref{table:white-cifar100}, we find AW-Net can achieve the best W-Robust Acc compared with state-of-the-art robust models with similar parameter scales. From the results of CIFAR-10 in Table \ref{table:white-cifar10}, compared to the best baseline models, AW-Net improves W-Robust Acc by 9.04\%, 1.57\%, 0.81\%, 0.35\%, and 0.43\% against FSGM, PGD$_{\scriptscriptstyle \rm{sat}}$, PGD$_{\scriptscriptstyle \rm{trades}}$, CW$_{\scriptscriptstyle \rm{\infty}}$, and AA attacks;  From the results of CIFAR-100 in Table \ref{table:white-cifar100}, compared to the best baseline models, AW-Net improves W-Robust Acc by 11.39\%, 2.92\%, 2.26\%, 1.47\%, and 0.92\% against FSGM, PGD$_{\scriptscriptstyle \rm{sat}}$, PGD$_{\scriptscriptstyle \rm{trades}}$, CW$_{\scriptscriptstyle \rm{\infty}}$, and AA attacks. These results verify the effectiveness of our AW-Net.

\setlength{\tabcolsep}{1pt}
\begin{table}[t]
\begin{center}
\caption{Black-box robustness on CIFAR-10 dataset. All the models are trained following one of the state-of-the-art adversarial training method MTARD \cite{Zhao2022Enhanced}.  All the results are the best checkpoints based on W-Robust Acc.}
\label{table:black-cifar10}
\scalebox{0.96}  { 
\begin{tabular}{
>{\columncolor[HTML]{FFFFFF}}c 
>{\columncolor[HTML]{FFFFFF}}c |
>{\columncolor[HTML]{FFFFFF}}c |
>{\columncolor[HTML]{FFFFFF}}c 
>{\columncolor[HTML]{FFFFFF}}c |
>{\columncolor[HTML]{FFFFFF}}c 
>{\columncolor[HTML]{FFFFFF}}c }
\hline 
% \noalign{\smallskip}
 &            & Clean               & \multicolumn{2}{c|}{\cellcolor[HTML]{FFFFFF}PGD$_{\scriptscriptstyle \rm{sat}}$} & \multicolumn{2}{c}{\cellcolor[HTML]{FFFFFF}CW$_{\scriptscriptstyle \rm{\infty}}$} \\ 
 % \noalign{\smallskip} 
 \hline 
 % \noalign{\smallskip}
Model                                                    & Params & $\mathcal{A}_{nat}$ & $\mathcal{A}_{adv}$      & $\mathcal{A}_{w}$     & $\mathcal{A}_{adv}$     & $\mathcal{A}_{w}$    \\ 
% \noalign{\smallskip} 
\hline \noalign{\smallskip}
ResNet-34\cite{he2016deep}              & 20.3M    & 90.30\%             & 67.67\%                  & 78.99\%               & 67.30\%                 & 78.80\%              \\
ResNet-50\cite{he2016deep}              & 22.6M    & 90.22\%             & 68.31\%                  & 79.27\%               & 68.12\%                 & 79.17\%              \\
WideResNet-34-8\cite{zagoruyko2016wide} & 29.5M    & 90.14\%             & 68.40\%                  & 79.27\%               & 67.53\%                 & 78.84\%              \\
VGG-16-BN\cite{simonyan2014very}        & 32.2M    & 86.89\%             & 65.93\%                  & 76.41\%               & 65.39\%                 & 76.14\%              \\
RepVGG-A2\cite{ding2021repvgg}          & 25.6M    & 81.05\%             & 62.62\%                  & 73.84\%               & 62.41\%                 & 71.73\%              \\
\textbf{AW-Net (ours)}                  & 26.3M    & 93.08\% &   66.93\% & \textbf{80.01\%} & 66.29\% & \textbf{79.69\%}          \\ \noalign{\smallskip} \hline
\end{tabular}
}
\end{center}
\end{table}
\setlength{\tabcolsep}{1.4pt}

\setlength{\tabcolsep}{1pt}
\begin{table}[t]
\begin{center}
\caption{Black-box robustness on CIFAR-100 dataset. All the models are trained following one of the state-of-the-art adversarial training method MTARD \cite{Zhao2022Enhanced}.  All the results are the best checkpoints based on W-Robust Acc.}
\label{table:black-cifar100}
\scalebox{0.96}  { 
\begin{tabular}{
>{\columncolor[HTML]{FFFFFF}}c 
>{\columncolor[HTML]{FFFFFF}}c |
>{\columncolor[HTML]{FFFFFF}}c |
>{\columncolor[HTML]{FFFFFF}}c 
>{\columncolor[HTML]{FFFFFF}}c |
>{\columncolor[HTML]{FFFFFF}}c 
>{\columncolor[HTML]{FFFFFF}}c }
\hline 
% \noalign{\smallskip}
&            & Clean               & \multicolumn{2}{c|}{\cellcolor[HTML]{FFFFFF}PGD$_{\scriptscriptstyle \rm{sat}}$} & \multicolumn{2}{c}{\cellcolor[HTML]{FFFFFF}CW$_{\scriptscriptstyle \rm{\infty}}$} \\ 
% \noalign{\smallskip}
\hline 
% \noalign{\smallskip}
Model                                   & Params & $\mathcal{A}_{nat}$ & $\mathcal{A}_{adv}$      & $\mathcal{A}_{w}$     & $\mathcal{A}_{adv}$     & $\mathcal{A}_{w}$    \\ 
% \noalign{\smallskip} 
\hline \noalign{\smallskip}
ResNet-34\cite{he2016deep}              & 20.3M    & 65.69\%             & 41.92\%                  & 53.81\%               & 42.10\%                 & 53.90\%              \\
ResNet-50\cite{he2016deep}              & 22.6M    & 69.59\%             & 45.13\%                  & 57.36\%               & 44.93\%                 & 57.26\%              \\
WideResNet-34-8\cite{zagoruyko2016wide} & 29.5M    & 60.46\%             & 38.81\%                  & 49.64\%               & 39.37\%                 & 49.92\%              \\
VGG-16-BN\cite{simonyan2014very}        & 32.2M    & 57.65\%             & 38.85\%                  & 48.25\%               & 39.21\%                 & 48.43\%              \\
RepVGG-A2\cite{ding2021repvgg}          & 25.6M    & 51.44\%             & 36.21\%                  & 43.83\%               & 37.08\%                 & 44.26\%              \\
\textbf{AW-Net (ours)}                  & 26.3M    & 73.98\% & 42.66\% & \textbf{58.32\%} & 42.93\% &\textbf{58.46\%}            \\  \noalign{\smallskip} \hline
\end{tabular}
}
\end{center}
\end{table}
\setlength{\tabcolsep}{1.4pt}

\noindent \textbf{Robustness against Black-box Attacks}.
Similar to other works \cite{wang2019improving,zhang2019theoretically,Zhao2022Enhanced}, we use the transfer-based attacks to evaluate the models' black-box robustness. For that, we conduct the white-box attacks on a surrogate model and then evaluate the black-box robustness using these generated adversarial examples. Here,  we select the strong robust model: WideResNet-70-16 \cite{gowal2020uncovering} as a surrogate model to generate adversarial examples. 

The results in Table \ref{table:black-cifar10} and Table \ref{table:black-cifar100} show that our AW-Net can outperform the state-of-the-art robust models and achieve better W-robust Acc. AW-Net improve W-robust Acc by 0.74\% and 0.52\% on CIFAR-10 against PGD$_{\scriptscriptstyle \rm{sat}}$ attack and CW$_{\scriptscriptstyle \rm{\infty}}$ attack respectively. Meanwhile, AW-Net can achieve similar performance on CIFAR-100, which has an improvement of 0.96\% and 1.20\% against the PGD$_{\scriptscriptstyle \rm{sat}}$ and CW$_{\scriptscriptstyle \rm{\infty}}$ attack. The robustness against black-box attacks shows the effectiveness of our AW-Net.

\subsubsection{Comparisons with other model architectures}
Besides the above static convolution neural networks, we choose two other kinds of popular network architectures to conduct comparisons. One is the well-known transformer architecture, and another is dynamic networks. 

\noindent \textbf{Comparisons with ViT networks:}  {To verify the effectiveness of our method, we select Swin Transformer \cite{liu2021swin} to compare with our method. The Swin Transformer is trained based on the \cite{mo2022adversarial}, which is specially designed for training robust ViT models. And the AW-Net is trained by \cite{Zhao2022Enhanced}. The result in Table \ref{table:vit} shows that our AW-Net achieves an improvement of 5.28\% and 3.5\% compared with Swin-Tiny and Swin-Small under the AA attack, even with smaller model scales (our model is 26.3M, which is smaller than Swin-Tiny's 27.5M and Swin-Small's 48.8M).}

\setlength{\tabcolsep}{1pt}
\begin{table}[ht]
\begin{center}
\caption{ {The comparison with ViT networks and other dynamic networks.  }}
\label{table:vit}
\scalebox{0.90}  { 
\begin{tabular}{
>{\columncolor[HTML]{FFFFFF}}c 
>{\columncolor[HTML]{FFFFFF}}c |
>{\columncolor[HTML]{FFFFFF}}c |
>{\columncolor[HTML]{FFFFFF}}c 
>{\columncolor[HTML]{FFFFFF}}c |
>{\columncolor[HTML]{FFFFFF}}c 
>{\columncolor[HTML]{FFFFFF}}c }
\hline 
% \noalign{\smallskip}
&           & Clean               & \multicolumn{2}{c|}{\cellcolor[HTML]{FFFFFF}PGD$_{\scriptscriptstyle \rm{sat}}$} & \multicolumn{2}{c}{\cellcolor[HTML]{FFFFFF}AA} \\ 
% \noalign{\smallskip}
\hline 
% \noalign{\smallskip}
Model                                   & Params & $\mathcal{A}_{nat}$ & $\mathcal{A}_{adv}$      & $\mathcal{A}_{w}$     & $\mathcal{A}_{adv}$     & $\mathcal{A}_{w}$    \\ 
% \noalign{\smallskip} 
\hline \noalign{\smallskip}

Swin-Tiny\cite{liu2021swin}   & 27.5M    & 80.71\%             & 49.79\%                  & 65.25\%               & 46.36\%                 & 63.54\%              \\
Swin-Small\cite{liu2021swin}  & 48.8M    & 84.46\%             & 50.02\%                  & 67.24\%               & 46.17\%                 &  65.32\%              \\

RDI-ResNet-38\cite{hu2020triple} & 2.7M  & 83.79\%            & 43.28\%                  & 63.54\%               &  -                        & -          \\
ADVMOE\cite{zhang2023robust} & 29.5M  & 84.32\%            & 55.73\%                  & 70.03\%               &  45.89\%                        & 65.11\%          \\

\textbf{AW-Net (ours)}        & 26.3M    & 93.08\%             & 50.93\%                  & \textbf{72.01\%}               & 44.56\%                 & \textbf{68.82\%}             \\  
%\textbf{AW-Net-L (ours)}     & 38.8M    & 92.50\%              & 52.74\%                  & \textbf{72.62\%}       & 45.78\%                & \textbf{69.14\%}              \\  
\noalign{\smallskip} \hline
\end{tabular}
}
\end{center}
\end{table}
\setlength{\tabcolsep}{1.4pt}

\noindent \textbf{Comparisons with dynamic networks:} Besides the static networks above, we here compare our AW-Net with a robust multi-exit dynamic weight network: RDI-ResNet-38 \cite{hu2020triple}  {and a robust mixture-of-expert network: ADVMOE \cite{zhang2023robust}.}  The evaluation is based on the PGD$_{\scriptscriptstyle \rm{sat}}$ and AA. The result in Table \ref{table:vit} shows that AW-Net has an obvious advantage compared with RDI-ResNet-38 or ADVMOE. Compared with ADVMOE, our AW-Net improves W-Robust Acc by 1.98\% and 3.71\% under the PGD$_{\scriptscriptstyle \rm{sat}}$ and AA attacks. This further indicates the effectiveness of the proposed dynamic network versus the varied weights. 
%\begin{table}[ht]
%\begin{center}
%\caption{The robustness comparison with other dynamic networks.}
%\label{Dynamic Networks}
%\begin{tabular}{m{1cm}<{\centering}m{2.8cm}<{\centering}|m{1.1cm}<{\centering}m{1.1cm}<{\centering}m{1.1cm}<{\centering}}
%\hline
% Attack& model  &$\mathcal{A}_{nat}$ &$\mathcal{A}_{adv}$& $\mathcal{A}_{w}$  \\
%\hline
%\multirow{2}*{PGD$_{\scriptscriptstyle \rm{sat}}$}& RDI-ResNet-38  \cite{hu2020triple} & 83.79\% & 43.28\% & 63.54\% \\
%~ & AW-Net (ours) & 93.08\% & 50.93\% & \textbf{72.01\%}  \\
%\hline
%\end{tabular}
%\end{center}
% \vspace{-0.5cm}
%\end{table}

 {At the same time, we also try to use some adversarial attack methods against dynamic weight networks to test our AW-Net. The robustness performance against DeepSloth \cite{hong2020panda} is 80.46\%, which shows the effectiveness against these types of attack methods. Actually, these attacks focus on the multi-exit dynamic weight network, while our AW-Net is an end-to-end dynamic weight network, which exists a fundamental gap in model structure and is also the reason why our AW-Net can defend against this type of method.  }

\subsubsection{Comparisons in RobustBench}

 {To verify the effectiveness of our method, we compare it with a state-of-the-art adversarial training method in RobustBench \cite{croce2021robustbench}: HAT \cite{rade2021helper}. We select network architectures including ResNet-18 \cite{he2016deep} and WideResNet-34-10 \cite{zagoruyko2016wide} used in HAT to compare with our method. All the methods are trained without additional generated data. We report the result of our AW-Net and AW-Net-L (Described in Section. \ref{Effects of joint design}). 
 % Since the parameter scale of our original AW-Net is quite smaller than WideResNet-34-10, we increase the parameter scale: we replace the basis of ResNet-18 used in the original backbone with a basis of ResNet-34, and we call this version AW-Net-L. 
The final results in Table \ref{table:other baseline} show that our AW-Net-L achieves better performance with a smaller parameter scale compared to the WideResNet-34-10 trained by HAT \cite{rade2021helper}. Under the attack of PGD$_{\scriptscriptstyle \rm{sat}}$ and AA, our AW-Net-L improves W-Robust Accuracy by 2.5\% and 0.2\% with less parameters (46.1M vs 38.8M), which verifies our superiority.}

\setlength{\tabcolsep}{1pt}
\begin{table}[ht]
\begin{center}
\caption{ {The comparison with SOTA adversarial training method HAT \cite{rade2021helper} in RobustBench. AW-Net-L denotes the larger version of our AW-Net. WRN-34-10 denotes the WideResNet-34-10.}}
\label{table:other baseline}
\scalebox{0.95}  { 
\begin{tabular}{
>{\columncolor[HTML]{FFFFFF}}c |
>{\columncolor[HTML]{FFFFFF}}c |
>{\columncolor[HTML]{FFFFFF}}c |
>{\columncolor[HTML]{FFFFFF}}c 
>{\columncolor[HTML]{FFFFFF}}c |
>{\columncolor[HTML]{FFFFFF}}c 
>{\columncolor[HTML]{FFFFFF}}c }
\hline 
% \noalign{\smallskip}
&      & Clean               & \multicolumn{2}{c|}{\cellcolor[HTML]{FFFFFF}PGD$_{\scriptscriptstyle \rm{sat}}$} & \multicolumn{2}{c}{\cellcolor[HTML]{FFFFFF}AA} \\ 
% \noalign{\smallskip}
\hline 
% \noalign{\smallskip}
Model  & Param      & $\mathcal{A}_{nat}$ & $\mathcal{A}_{adv}$      & $\mathcal{A}_{w}$     & $\mathcal{A}_{adv}$     & $\mathcal{A}_{w}$    \\ 
% \noalign{\smallskip}
\hline \noalign{\smallskip}
% WRN-34-10\cite{zagoruyko2016wide} & \cite{wang2019improving}        & 46.1M    & 84.17\%             & 58.56\%                  & 71.37\%               & 51.10\%                 & 67.74\%               \\
% WRN-34-10\cite{zagoruyko2016wide} & \cite{zhang2019theoretically} & 46.1M    & 84.91\%             & 55.30\%                  & 70.11\%               & 53.07\%                 & 68.99\%              \\
ResNet-18 + HAT & 10.7M    &   84.90\% & 52.02\% & 68.46\% & 49.08\% & 66.99\%          \\
WRN-34-10 + HAT & 46.1M    &  86.12\% & 54.12\% & 70.12\% & 51.75\% & 68.94\%            \\
\textbf{AW-Net (ours)}        & 26.3M    & 93.08\%             & 50.93\%                  & 72.01\%               & 44.56\%                 & 68.82\%             \\  
\textbf{AW-Net-L (ours)}              & 38.8M    &   92.50\% & 52.74\% & \textbf{72.62\%} & 45.78\% & \textbf{69.14\%}               \\  \noalign{\smallskip} \hline
\end{tabular}
}
\end{center}
\end{table}
\setlength{\tabcolsep}{1.4pt}

\subsubsection{Robustness against adaptive attacks}
According to the adaptive attack definition in \cite{tramer2020adaptive}, we augment the original objective function used in the {PGD$_{\scriptscriptstyle \rm{trades}}$} attack with our joint training loss (i.e., Eq.(\ref{eq12})) to implement adaptive attacks. The comparison results of CIFAR-10 and CIFAR-100 under different attacks are shown in Table \ref{adaptive}.
 \setlength{\tabcolsep}{1pt}
\begin{table}[h]\scriptsize
\begin{center}
\caption{Adaptive Attack for AW-Net. The attack is based on {PGD$_{\scriptscriptstyle \rm{trades}}$}.}
\label{adaptive}
\begin{tabular}{m{1.5cm}<{\centering}m{3cm}<{\centering}|m{1.5cm}<{\centering}}
\hline
 Dataset & Attacks  &$\mathcal{A}_{adv}$  \\
\hline
\multirow{2}*{CIFAR-10} & {PGD$_{\scriptscriptstyle \rm{trades}}$}  &52.37\% \\
~ & Adaptive attack  & 54.07\%  \\
\hline
\multirow{2}*{CIFAR-100}  &{PGD$_{\scriptscriptstyle \rm{trades}}$}  &27.51\%  \\
~ & Adaptive attack & 27.93\%  \\
\hline
\end{tabular}
\end{center} 
\vspace{-0.3cm}
\end{table}
\setlength{\tabcolsep}{1.4pt}
Under the adaptive attack, we see that the Robust Accuracy of CIFAR-10 is 54.07\%, which is similar to the Robust Accuracy under the original {PGD$_{\scriptscriptstyle \rm{trades}}$} attack (52.37\%). In the CIFAR-100 dataset, the Robust Accuracy under adaptive attack is also similar to the original attack (27.51\% vs 27.93\%). These results verify the robustness of AW-Net against adaptive attacks.

\subsection{Robustness on Large-scale Dataset}
Besides CIFAR-10 and CIFAR-100 datasets, we also verify our method on the Tiny-ImageNet, a subset of the well-known ImageNet.
For the Tiny-ImageNet dataset, we compare it with ResNet-34 \cite{he2016deep} and ResNet-50 \cite{he2016deep}. In table 1 and Table 2, ResNet-50 shows the second best performance, and thus we select it as the main comparison model here. 
Both all the models are trained following MTARD \cite{Zhao2022Enhanced}.
We train the above models for 50 epochs, and the learning rate is divided by 25-th and 40-th epochs. The batch size is 128 and $\tau$ is set to 5. $\beta$ of AW-Net is set to 3. 
% Both our AW-Net and the compared ResNet are trained with state-of-the-art adversarial training algorithm: MTARD \cite{Zhao2022Enhanced}. 
% We train the above models for 50 epochs, and the learning rate is divided by 25-th and 40-th epochs. The batch size is 128 and $\tau$ is set to 5. $\beta$ of AW-Net is set to 3. 

\setlength{\tabcolsep}{1pt}
\begin{table}[ht]\scriptsize
\begin{center}
\caption{White-box Robustness on Tiny-ImageNet. All the results are the best checkpoints based on $\mathcal{A}_{w}$. }
\label{table:5}
\begin{tabular}{m{1.2cm}<{\centering}m{2.6cm}<{\centering}|m{1.1cm}<{\centering}m{1.1cm}<{\centering}m{1.1cm}<{\centering}}
\hline
Attack & Models &$\mathcal{A}_{nat}$ &$\mathcal{A}_{adv}$& $\mathcal{A}_{w}$  \\
\hline
\multirow{3}*{FGSM} & ResNet-34 & 47.00\%  & 41.30\% & 44.15\% \\
& ResNet-50 & 61.08\% & 23.68\% & 42.38\% \\
~ & \textbf{AW-Net(ours)} & 63.32\% & 43.32\% & \textbf{53.32\%}  \\
\hline
\multirow{3}*{PGD$_{\scriptscriptstyle \rm{sat}}$}  & ResNet-34 & 47.00\% & 22.70\% & 34.85\% \\
& ResNet-50 & 61.08\% & 16.25\% & 38.67\% \\
~ & \textbf{AW-Net(ours)} & 63.32\% & 17.94\% & \textbf{40.63\%} \\
\hline
\multirow{3}*{PGD$_{\scriptscriptstyle \rm{trades}}$}  &  ResNet-34 & 47.00\% & 23.83\% & 35.42\% \\
& ResNet-50 & 61.08\% & 18.11\% & 39.60\% \\
~ & \textbf{AW-Net(ours)} & 63.32\% & 19.43\% & \textbf{41.38\%} \\
\hline
\multirow{3}*{CW$_{\scriptscriptstyle \rm{\infty}}$}  & ResNet-34 & 47.00\% & 20.94\% & 33.97\% \\
& ResNet-50 & 61.08\% & 14.82\% & 37.95\% \\
~ & \textbf{AW-Net(ours)} & 63.32\% & 13.67\% & \textbf{38.50\%}  \\
\hline
\multirow{3}*{AA}  & ResNet-34 & 47.00\% & 18.64\% & 32.82\% \\
& ResNet-50 & 61.08\% & 10.29\% & 35.69\% \\
~ & \textbf{AW-Net(ours)} & 63.32\% & 8.61\% & \textbf{35.97\%}  \\
\hline
\end{tabular}
\end{center} 
% \vspace{-0.5cm}
\end{table}
\setlength{\tabcolsep}{1.4pt}

The results on the Tiny-ImageNet are given in Table \ref{table:5}, showing that AW-Net can outperform the ResNet-34 and ResNet-50 when facing different attacks. Specifically, compared to the best baseline models, AW-Net improves W-Robust Acc by 9.17\%, 1.96\%, 1.78\%, 0.55\%, and 0.28\% against FSGM, PGD$_{\scriptscriptstyle \rm{sat}}$, PGD$_{\scriptscriptstyle \rm{trades}}$, CW$_{\scriptscriptstyle \rm{\infty}}$, and AA attacks. The results show that AW-Net can achieve better performance and is still effective in a more challenging dataset.

\section{Conclusion}
\label{conclusion}
In this paper, we theoretically explored the difference between standard-trained and robust-trained models versus the filters' weight distribution for the same network and argued that static neural networks were hard to address the trade-off issue. Then we proposed the Adversarial Weight-Varied Network (AW-Net) to deal with the clean and adversarial example with different filters' weights to self-adapt input samples. We used MixBN to fit the distributions of clean and adversarial examples and applied an adversarial router sub-network to distinguish the clean and adversarial samples. The generated regulation signals were applied to guide the adjustment of the dynamic network. A series of solid experiments proved that AW-Net could achieve state-of-the-art performance in W-Robust accuracy compared with other robust models on three datasets.

% \section*{Data Availability Statements}
% The datasets generated during and/or analysed during the current study are available from the corresponding author on reasonable request.

\ifCLASSOPTIONcaptionsoff
  \newpage
\fi

\bibliographystyle{splncs04}
\bibliography{egbib}

\vspace{-1cm}
\begin{IEEEbiography}[{\includegraphics[width=1in,height=1.25in,clip,keepaspectratio]{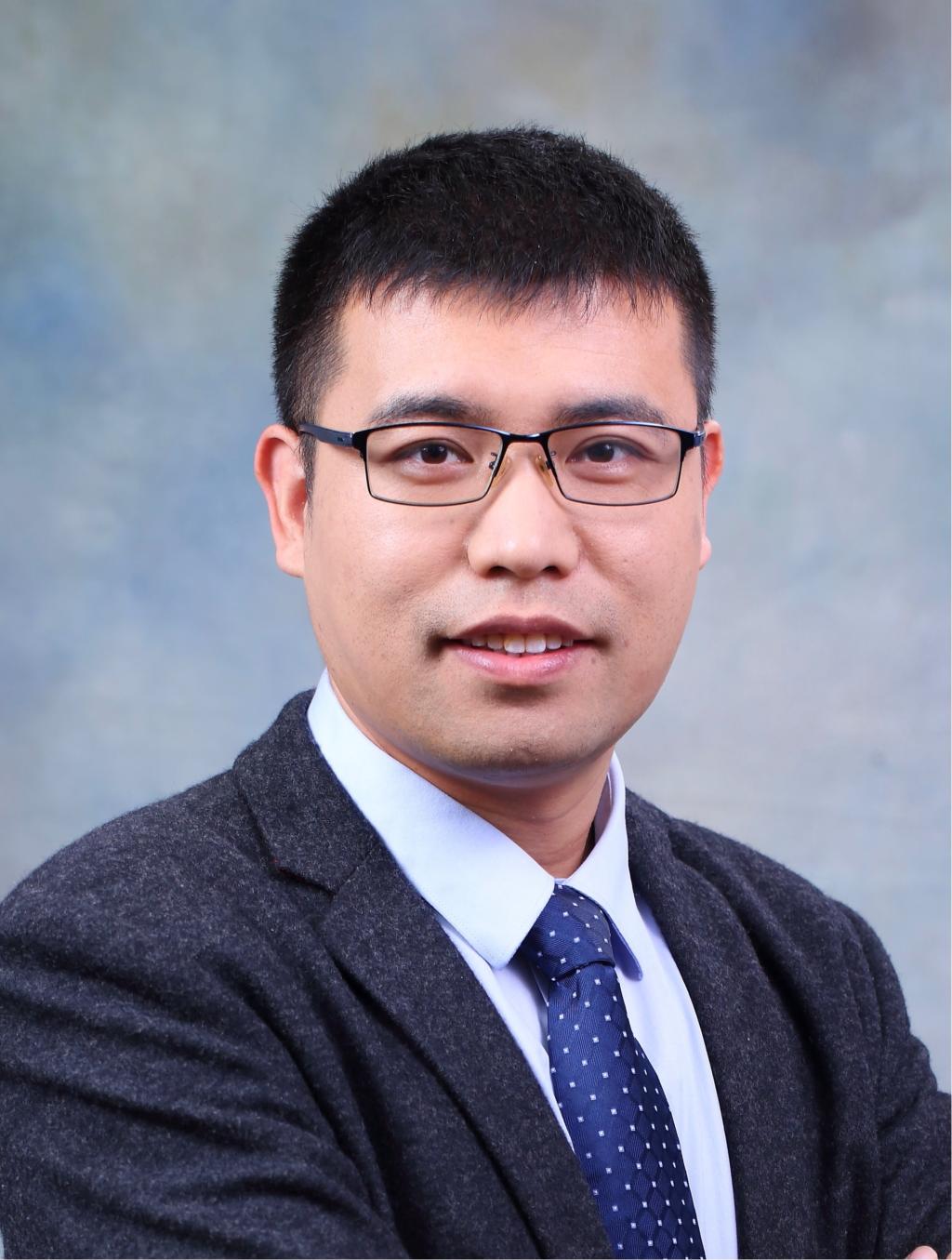}}]{Xingxing Wei} received his Ph.D degree in computer science from Tianjin University, and B.S.degree in Automation from Beihang University (BUAA), China. He is now an Associate Professor at Beihang University (BUAA). His research interests include computer vision, adversarial machine learning and its applications to multimedia content analysis. He is the author of referred journals and conferences in IEEE TPAMI, TMM, TCYB, TGRS, IJCV, PR, CVIU, CVPR, ICCV, ECCV, ACMMM, AAAI, IJCAI etc.
\end{IEEEbiography}
\vspace{-1cm}
\begin{IEEEbiography}[{\includegraphics[width=1in,height=1.25in,clip,keepaspectratio]{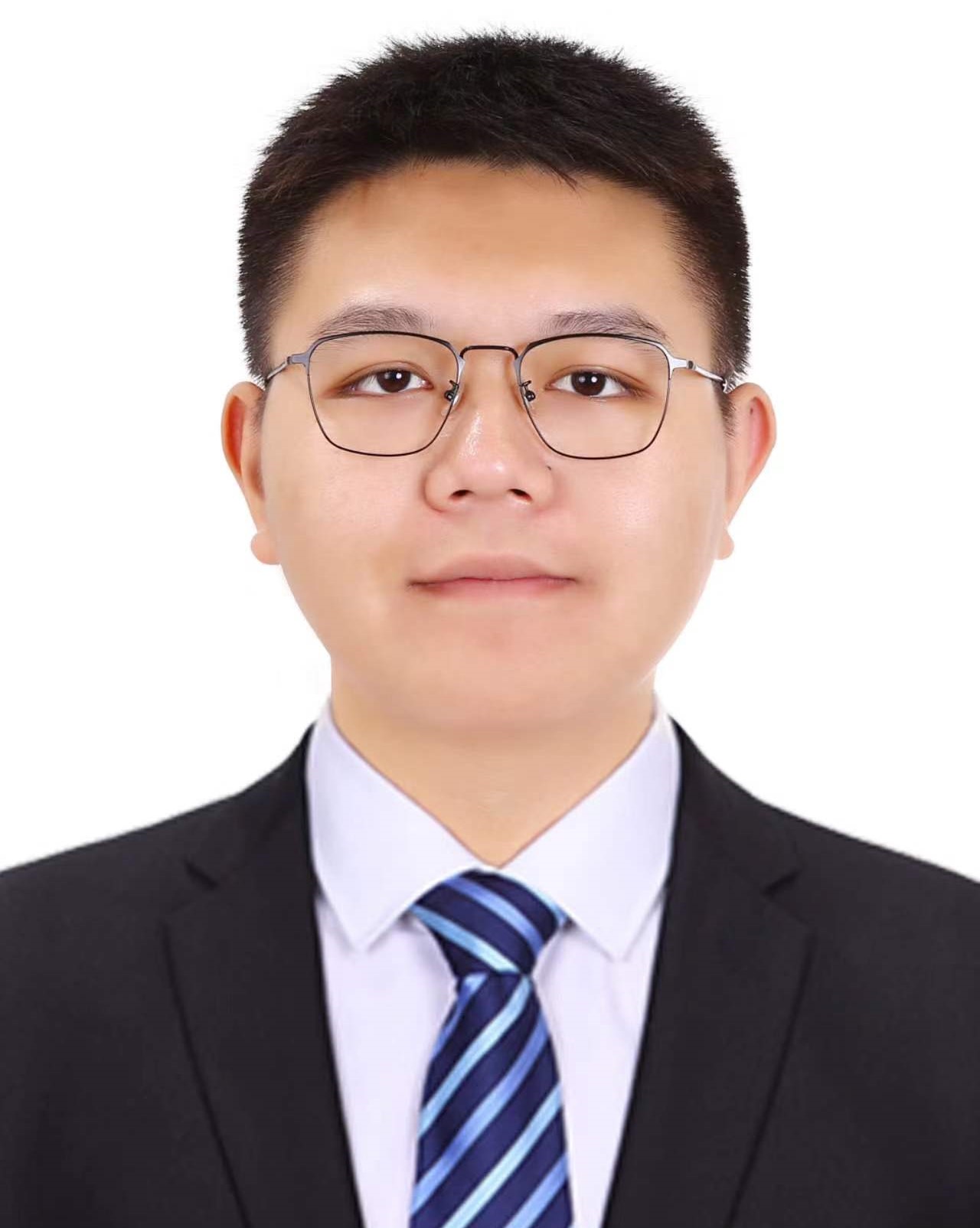}}]{Shiji Zhao} received his B.S. degree in the School of Computer Science and Engineering, Beihang University (BUAA), China. He is now a Ph.D student in the Institute of Artificial Intelligence, Beihang University (BUAA), China. His research interests include computer vision, deep learning and adversarial robustness in machine learning.
\end{IEEEbiography}
\vspace{-1cm}
\begin{IEEEbiography}[{\includegraphics[width=1.4in,height=1.25in,clip,keepaspectratio]{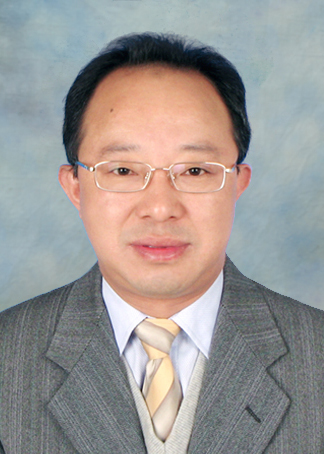}}]{Bo Li} Bo Li is currently a Changjiang distinguished professor of the School of Computer Science and Engineering, Beihang University. He is a recipient of The National Science Fund for Distinguished Young Scholars. He is currently the dean of AI Research Institute, Beihang University. He is the chief scientist of National 973 Program and the Principal investigator of the National Key Research and Development Program. He has published more than 100 papers in top journals and conferences.
\end{IEEEbiography}

\end{document}